\documentclass[final,twocolumn]{svjour3}

\usepackage{times}
\usepackage{epsfig}
\usepackage{graphicx}
\usepackage{amsmath}
\usepackage{amssymb}
\usepackage{caption}

\usepackage{comment}
\usepackage{algorithm}
\usepackage[algo2e,vlined,algoruled,resetcount]{algorithm2e}
\usepackage{enumitem}
\usepackage{multirow}

\usepackage{subfig}
\usepackage{eucal}

\usepackage{cite}

\usepackage[numbers]{natbib}

\usepackage[pagebackref=true,breaklinks=true,letterpaper=true,colorlinks,bookmarks=false,citecolor=red]{hyperref}
\hypersetup{linkcolor=red}

\usepackage{cases}

\newcommand{\be}{\mathbf e}
\newcommand{\bu}{\mathbf u}
\newcommand{\bx}{\mathbf x}

\newcommand{\by}{\mathbf y}

\newcommand{\bv}{\mathbf v}
\newcommand{\bb}{\mathbf b}

\newcommand{\bw}{\mathbf w}

\newcommand{\bX}{\mathbf X}

\newcommand{\bP}{\mathbf P}

\newcommand{\bW}{\mathbf W}
\newcommand{\bA}{\mathbf A}

\newcommand{\bH}{\mathbf H}
\newcommand{\bC}{\mathbf C}

\newcommand{\bB}{\mathbf B}
\newcommand{\bZ}{\mathbf Z}
\newcommand{\bY}{\mathbf Y}

\newcommand{\fE}{\mathrm{E}}

\newcommand{\fd}{\mathrm{d}}
\newcommand{\ftheta}{\mathrm{\theta}}
\newcommand{\frank}{\mathrm{rank}}

\newcommand{\fbound}{\mathrm{bound}}
\newcommand{\fbranch}{\mathrm{split}}
\newcommand{\ftrace}{\mathrm{trace}}

\newcommand{\setD}{\mathcal{D}}
\newcommand{\setV}{\mathcal{V}}
\newcommand{\setE}{\mathcal{E}}
\newcommand{\setF}{\mathcal{F}}

\newcommand{\setZ}{\mathcal{Z}}
\newcommand{\setI}{\mathcal{I}}
\newcommand{\setJ}{\mathcal{J}}
\newcommand{\setJbar}{\bar{\setJ}}
\newcommand{\setQ}{\mathcal{Q}}
\newcommand{\setG}{\mathcal{G}}

\newcommand{\real}{\mathbb{R}}
\newcommand{\psdd}{\succcurlyeq}

\newcommand{\sst}{\mathrm{s.t.}}

\newcommand{\GLB}{\mathrm{glb}}
\newcommand{\GUB}{\mathrm{gub}}
\newcommand{\LB}{\mathrm{lb}}
\newcommand{\UB}{\mathrm{ub}}

\def\T{{\!\top}}

\newcommand{\tabincell}[2]{\begin{tabular}{@{}#1@{}}#2\end{tabular}}

\def\psd{p.s.d\onedot}

\ifx\theorem\undefined
\newtheorem{theorem}{Theorem}
\newenvironment{theorem*}{\par\noindent{\bf Theorem\ }}{\hfill\\[2mm]}

\newtheorem{proposition}[theorem]{Proposition}

\newenvironment{corollary*}{\par\noindent{\bf Corollary\ }}{\hfill\\[2mm]}

\fi

\usepackage{marvosym}

\def\BC{{B\&C}\xspace}
\def\BB{{B\&B}\xspace}

\usepackage{graphicx}
\def\tvdots{\raisebox{3pt}{$\scalebox{.75}{\vdots}$}}

\expandafter\def\expandafter\normalsize\expandafter{%
\normalsize\setlength\abovedisplayskip{3pt}}

\expandafter\def\expandafter\normalsize\expandafter{%
\normalsize\setlength\belowdisplayskip{3pt}}

\makeatletter
\DeclareRobustCommand\onedot{\futurelet\@let@token\@onedot}
\def\@onedot{\ifx\@let@token.\else.\null\fi\xspace}

\def\eg{{for example\xspace}} 
\def\ie{\emph{i.e}\onedot} 
 
 \def\vs{\emph{vs}\onedot}
\def\wrt{with respect to\xspace} 
\def\etal{\emph{et al}\onedot}
\makeatother

\usepackage{color}
\definecolor{blue}{rgb}{1,1,1}

\journalname{manuscript}
\begin{document}

\title{Efficient Semidefinite Branch-and-Cut for MAP-MRF Inference }

\author{Peng Wang \and Chunhua Shen
                  \and Anton van den Hengel
                  \and Philip H. S. Torr}
\institute{
   P. Wang
   \at  University of Adelaide, SA 5005, Australia.
   \and
   C. Shen (\Letter)
   \at  University of Adelaide, SA 5005, Australia;
   \at and Australian Centre for Robotic Vision.
   \\
   \email{chunhua.shen@adelaide.edu.au}
   \and
   A. van den Hengel
   \at  University of Adelaide, SA 5005, Australia;
   \at
   and Australian Centre for Robotic Vision.
   \and
   P. H. S. Torr
   \at  University of Oxford, United Kingdom.
}

\maketitle
\begin{abstract}

We propose a Branch-and-Cut  (\BC) method for solving general MAP-MRF inference problems.
  { The core of our method is a very efficient bounding procedure,
  which combines scalable semidefinite programming (SDP) and a cutting-plane method
  for seeking violated constraints.
  In order to further speed up the computation, several strategies have been exploited,
  including model reduction, warm start and removal of inactive constraints.}

We analyze the performance of the proposed method under different settings,
and demonstrate that
our method either outperforms or performs on par with state-of-the-art approaches.
Especially when the connectivities are dense or when the
relative magnitudes of the unary costs are low, we achieve the best reported results.
Experiments show that the proposed algorithm achieves better approximation
than the state-of-the-art methods within a variety of time budgets on
challenging non-submodular MAP-MRF inference problems.

\end{abstract}

\section{Introduction}

Markov Random Fields (MRFs) have been used to model a variety of problems in
computer vision, including semantic image segmentation, restoration, 3D reconstruction and stereo matching, amongst
a lot of others.
Finding the maximum a posteriori (MAP) solution to \emph{general} MRF problems is typically NP-hard, however,
and many approaches have been proposed to solve such problems, approximately or exactly
(see \cite{kappes2013comparative,szeliski2008comparative,Kumar2009Ananalysis,kolmogorov2006comparison,Givry2014exp} for comparative studies).

Graph cuts based methods~\cite{boykov2001fast,kolmogorov2004energy,kolmogorov2007minimizing}
have been applied to a number of MAP problems in computer vision.
Binary MRFs with submodular pairwise potentials can be solved exactly and efficiently by graph cuts.
The QPBO~\cite{rother2007optimizing, kolmogorov2007minimizing} algorithm can obtain a
part of the globally optimal solution for some non-submodular, binary, pairwise MRFs, but
its performance degrades as the portion of non-submodular potentials increases
and in the case of
highly connected graphs or weak unary potentials~\cite{rother2007optimizing}.
{
For multi-label MRFs, expansion move and
swap move algorithms \cite{boykov2001fast} have been employed to find a strong local optimum
with the property that no expansion move (swap move) can decrease the energy.
However, this optimality is guaranteed only if the binary subproblem at each iteration
is solved globally.
In particular, the binary subproblem is typically required to be submodular
to guarantee this optimality.
}

Another class of popular inference approaches is based on message passing.
Max-product belief propagation~\cite{pearl1988probabilistic,weiss2001optimality,sun2002stereo,felzenszwalb2006efficient}
obtains exact solutions to tree-structured graphs,
single-cycle graphs~\cite{aji1998convergence,weiss2000correctness},
and maximum weight matching on bipartite graphs~\cite{bayati2005maximum}.
For graphs with cycles, approximate solutions can also be obtained using max-product belief propagation,
but there is no convergence guarantee.
Tree-reweighted max-product (TRW) message-passing~\cite{wainwright2005map,kolmogorov2006convergent,kolmogorovoptimality}
differs from belief propagation
in that it maintains a lower-bound to the minimum energy and can be used to measure the quality of approximate solutions.
TRW message passing is proved~\cite{kolmogorovoptimality} to be exact for binary submodular MRFs.
Kolmogorov~\cite{kolmogorov2006convergent} proposed an improved version of TRW, called TRW-S,
with convergence guarantee that TRW does not have.

Many aforementioned approaches have underlying connections to
the standard linear programming (LP) relaxation optimizing over {\em
local marginal polytope} \cite{shlezinger1976syntactic,werner2007linear,wainwright2008graphical}.
The ordinary max-product updates~\cite{pearl1988probabilistic}
can be considered as a Lagrangian approach to the dual of the standard LP relaxation, for tree-structured graphs.
However, this relationship cannot be generalized to graphs with cycles.
For binary pairwise graphs, QPBO and TRW message passing
exactly solve the standard LP relaxation\ \cite{kolmogorovoptimality}.
The standard LP relaxation can achieve
exact solutions for tree-structured graphs
or binary graphs with submodular potentials.
It is proved in \cite{Kumar2009Ananalysis} that the standard LP relaxation
dominates second-order cone prog\-ra\-mming (SOCP) relaxation\ \cite{kumar2006solving}
and quadratic prog\-ramming (QP) relaxation\ \cite{ravikumar2006quadratic}.

Standard LP algorithms such as interior-point methods are usually computationally inefficient for large-scale MAP problems,
as they do not exploit the graph structure.
To this end, a number of specific approaches to solve LP
relaxation for MAP problems have been proposed,
such as block coordinate descent methods\
\cite{kolmogorov2006convergent,Globerson2007},
subgradient descent methods\ \cite{komodakis2011mrf,kappes2010mrf,jojic2010accelerated},
bundle methods\ \cite{kappes2012bundle},
proximal methods\ \cite{ravikumar2010message},
ADMM\ \cite{martins2011augmenting,aguiar2011augmented,meshi2011alternating}
and smoothing methods~\cite{hazan2010norm,johnson2007lagrangian,ravikumar2010message,savchynskyy2012efficient}.

However, it has been shown in
\cite{komodakis2008beyond,werner2008high,sontag2008,batra2011tighter}
that the standard LP relaxation is still not tight enough for
many hard MAP problems in real applications.
Especially, the above-mentioned methods usually perform poorly
for the class of densely-connected graphs with weak unary potentials and a large portion of non-submodular pairwise potentials
({\em which are of major interests in this paper}).
Because there are a large number of (potentially frustrated) long cycles in de\-n\-se\-ly-conn\-ect\-ed graphs,
the standard LP relaxation is likely to provide loose bounds.
Two approaches can be adopted to alleviate this issue.
The first direction is based on LP relaxation with high-order consistency constraints\
\cite{sontag2008,werner2008high,komodakis2008beyond,johnson2008convex,batra2011tighter,sontag2012efficiently};
and the second one is to use semidefinite programming (SDP) relaxation\
\cite{torr2003solving,jordan2004semidefinite,Olsson07solvinglarge,
Schellewald05,Joulin2010dis,peng2012approximate,peng2013cpvr,guibasscalable,NIPS2014_5341}.

As the standard LP relaxation only enforces edge consistency
constraints over pseudo-marginals of variables, it can be loose due to
the existence of violated high-order consistency constraints.
Cluster based methods~\cite{sontag2008,komodakis2008beyond,batra2011tighter,sontag2012efficiently}
can be used
to tighten the standard LP relaxation by incrementally adding high-order consistency constraints.
Methods for searching violated high-order constraints include
dictionary enumeration for triplets or other short cycles\
\cite{sontag2008,batra2011tighter}
and separation algorithms for long cycles\ \cite{sontag2007new,sontag2012efficiently}.

SDP relaxation provides an alternative tighter bound for
MAP-MRF inference problems, compared with LP relaxation.
SDP is of great importance in developing approximation algorithms for
some NP-hard optimization problems, and usually provides more accurate solutions.
In particular for the classical maxcut problem, it achieves the best
approximation ratio of $0.879$\ \cite{Goemans95improved}.
Primal-dual interior-point methods\ \cite{alizadeh1998primal,andersen2003implementing,nesterov1998primal,ye1994nl,Helmbergphd}
are considered as state-of-the-art general SDP solvers,
which in the worst case require $\mathcal{O}(m^3+mn^3h^3+m^2n^2h^2)$ arithmetic operations and $\mathcal{O}(m^2+n^2h^2)$
memory requirement
to solve SDP relaxation to an MAP problem with $n$ nodes, $h$ states per node and $m$ linear constraints.
The exponent in the polynomial complexity bound is
so great as to preclude  practical applications of interior-point methods to even
medium sized problems.
{\em
This has significantly hampered the use of SDP relaxations in MAP-MRF
inference.}

Some approaches~\cite{burer2003nonlinear,Joulin2010dis,peng2012approximate,NIPS2014_5341} solve SDP relaxation efficiently
based on the low-rank approximation of \psd matrix: $\bX = \bY \bY^\T$
(such that \psd constraints are eliminated).
However, these methods need to solve a sequence of non-convex problems,
and may get stuck in a local-optimal point.
Multivariate weight updates based
methods\ \cite{arora2012multiplicative,arora2005fast,arora2007combinatorial,hazan2008sparse,garber2011approximating,laue2012hybrid}
solve specific SDP problems inexactly, but need many iterations to
converge to accurate solutions.
Augmented Lagrangian methods\
\cite{malick2009regularization,zhao2010newton,wen2010alternating,guibasscalable}
have also been proposed for solving SDP
problems, which can be viewed as an instance of gradient descent methods\ \cite{rockafellar1973dual} and may have a slow
convergence rate.

It is shown in \cite{wainwright2008graphical} that MAP LP relaxation with local consistency constraints
and conventional SDP relaxation without considering local consistency constraints
are mutually incomparable,
which means that neither of them is tighter than the other.
Thus combining conventional SDP relaxation and standard/high-order local consistency constraints
together would produce an even tighter bound.
Unfortunately, this combination results in a very challenging optimization problem,
not only because of the compu\-ta\-t\-ion\-al inefficiency of SDP itself,
but also due to the large number of linear constraints arising from
standard/high-order LP relaxation.

In this paper, we propose an efficient SDP approximation/bounding approach,
which is faster than state-of-the-art competing methods and able to
handle a large number of linear constraints.
A Branch-and-Cut (\BC)
method is developed based on this SDP bounding procedure.
Our approach can be used to achieve either the exact solution or an
accurate approximation to general MAP-MRF inference problems with a time budget,
and does so at a lower computational cost than state-of-the-art competing methods.
The main contributions of this paper are as follows.
\begin{itemize}
  \item[$\bullet$]
 We present an approximate inference approach based on a scalable SDP algorithm~\cite{peng2013cpvr} and cutting-plane.
    The proposed formulation minimizes the energy
    over the intersection of the semidefinite and polyhedral
    outer-b\-ou\-n\-d\-s arising from SDP and standard/high-order LP
    relaxation respectively.
    Such  optimization schemes result in  SDP problems with a large
    number of linear constraints $m$.
    Significantly, our proposed SDP method  scales {\em linearly} in $m$ (i.e., $ {\cal O}(m) $)---a sharp
    contrast to
  standard interior-point methods with complexity of $ {\cal O}(m^3) $.
    The method proceeds by incrementally adding violated constraints until the required solution
    quality  is achieved.
    Our method is much faster than interior-point methods while still maintains a
    comparable lower-bound of the minimum energy.

  \item[$\bullet$]

 { The proposed SDP approximate/bounding method
   is embedded into a \BC framework for exactly solving MAP-MRF inference problems.
    We also introduce a few techniques  to optimize the bounding and branching procedures,
    including model reduction, warm start and removal of inactive constraints.}

\item[$\bullet$]

  We analyze the performance on a variety
of MRF graphic models, which demonstrates that our method performs better than
 recent competing approaches, especially
 when the graph connectivity is dense and/or the unary
potentials are weak.
\end{itemize}

\noindent
{\bf Related work }
We review some research work that is closely related to ours.
The proposed approach is motivated by the line of our prior work in \cite{Shen2011scalabledual,peng2013cpvr}.
While the work in
\cite{Shen2011scalabledual,peng2013cpvr}
focuses on distance metric learning and image segmentation,
our work here considers approximate and exact methods for general MAP-MRF inference problems.

Different \BB algorithms have been proposed for MAP-MRF inference.
Sun \etal~\cite{Sun2012MRFBB2} proposed a \BB method (refer to as MPLP-BB) based on LP relaxation.
DAOOPT~\cite{otten2012anytime} is another \BB method based on LP relaxation, which combines several sophisticated techniques,
including
And/Or s\-e\-a\-r\-ch sp\-a\-ce\-s, the mini-bucket approach and stochastic local search.
In our experiments, MPLP-BB and DAOOPT are outperformed by our method.
Peng~\etal \ \cite{peng2012approximate}
developed an approximate inference method for
optimizing over the intersection of semidefinite bound and local marginal polytope.
Note that the optimization technique in their work is  different from ours.
There, the \psd constraint and local marginal polytope are separated using dual decomposition,
and then the SDP subproblem is solved using non-convex QP, which results in a local optimum.
On the contrary, our method processes the \psd and linear constraints
altogether and
optimizes a convex problem with a guarantee of being globally optimal.

{
SDP-based \BB or \BC approaches have also been widely studied for integer/continuous quadratic problems.
Krislock \etal~\cite{krislock2012improved} introduced an SDP-based branch-and-bound (\BB) method to solve Max-Cut problems.
An experimental comparison was performed in \cite{armbruster2012lp}
between LP- and SDP-based \BB methods for graph bisection problems,
which showed that SDP-based methods are better for graphs with a few thousands nodes.
The algorithm in \cite{burer2008finite} solves {\em real-valued} nonconvex quadratic programs
using a finite \BB algorithm in which the bound of each branch is computed by semidefinite relaxation.
The work in \cite{dinh2010efficient} introduced an exact solver for binary quadratic programs,
which combines DC (difference of convex functions) algorithms and SDP-based \BB approaches.
Mars and Schewe~\cite{mars2012sdp} proposed an SDP-based \BB method for mixed-integer semidefinite programs.
{ SDP-based cutting plane algorithms were studied in
\cite{helmberg1998cutting,helmberg2000semidefinite,helmberg1996quadratic,helmberg1998solving,helmberg1995combining} for binary quadratic programs,
in which linear constraints arising from cut polytope were used to strengthen the SDP relaxation.}
In contrast to the above approaches,
our algorithm aims to solve the MAP inference problems for discrete graphical models,
and uses a number of specific speedup strategies.
}

\noindent
{\bf Notation}
 Table~\ref{tab:notation}  lists the notation used in this paper:
\begin{table}[h]
  \centering
  \footnotesize
  \vspace{-0.25cm}
  \begin{tabular}{@{\hspace{0.1cm}}r@{\hspace{0.3cm}} p{5.8cm}}
  \hline
     $\bX$                            &A matrix (bold upper-case letters). \\
     $\bx$                            &A column vector (bold lower-case letters). \\
     $\mathbb{R}$                     &The set of real numbers. \\
     $\mathcal{S}^n$                  &The set of $n \times n$ symmetric matrices. \\
     $\mathcal{S}^n_+$                &The cone of positive semidefinite (\psd) $n \times n$ matrices. \\
     $\bX \psdd 0$                    &The matrix $\bX$ is positive semidefinite.\\
     $\leq, \geq$                     &Inequality between scalars or element-wise inequality between column vectors. \\
     $\mathbf{1}(\cdot)$              &Indicator function, $1$ if the statement is true and $0$ otherwise. \\
     $\ftrace(\cdot)$                 &Trace of a matrix. \\
     $\frank(\cdot)$                  &Rank of a matrix. \\
     $\mathrm{diag}(\bX)$             &The main diagonal vector of the matrix $\bX$. \\
     $\mathrm{diag}(\bx)$             &A diagonal matrix consisting of the input vector $\bx$ as diagonal elements. \\
     $\lVert \cdot \rVert_1$, $\lVert \cdot \rVert_2$   &$\ell_1$ and $\ell_2$ norm of a vector. \\
     $\lVert \cdot \rVert_F$        & Frobenius-norm of a matrix. \\
     $\langle \cdot, \cdot \rangle$   & Inner product of two matrices. \\
     $\nabla \mathrm{f}(\cdot)$       & The first-order derivatives of function $\mathrm{f}(\cdot)$.\\
     $\lfloor x \rfloor$              & The nearest integer less than or equal to $x$. \\
  \hline
  \end{tabular}
\caption{Notation.}
\label{tab:notation}
\end{table}

The remaining of this paper is organized as follows.
We first introduce the basic concepts of LP and SDP relaxation to MAP problems
in Section~\ref{sec:lp_sdp_intro}.
Then our main contributions,
the overall branch-and-cut method for energy minimization
and the SDP bounding approach with cutting-plane are discussed in
Sections~\ref{sec:bcsdp} and \ref{sec:sdp_relaxation} respectively.
The performance of our algorithms is evaluated experimentally in Section~\ref{sec:exp}.

\section{LP and SDP Relaxation to MAP-MRF Inference Problems}
\label{sec:lp_sdp_intro}

Suppose that an MAP-MRF inference model is characterized by an undirected graph $\setG = (\setV, \setE)$,
where $\setV$ represents the set of $n$ nodes (which is reused to represent the set of node indexes$\{1, 2, \dots, n \}$)
and $\setE$ represents the set of edges.
In this paper, we only consider unary and pairwise potentials,
and so MAP-MRF inference problems can be expressed as the following energy minimization problem:
\begin{align}
\min_{\bx \in \setZ^n} \fE (\bx) := \sum_{p \in \setV} \ftheta_p (x_p) + \sum_{p < q,(p,q) \in \setE} \ftheta_{p,q} (x_p, x_q),
\label{eq:energy0}
\end{align}
where $\ftheta_p(x_p)$ and $\ftheta_{p,q} (x_p, x_q)$ represent the unary and pairwise potentials respectively.
$\setZ = \{1,2,\dots,h\}$ refers to the set of $h$ possible states.
Without loss of generality, we assume that
all nodes have the same number of possible states.

It is well known that
the energy minimization problem \eqref{eq:energy0} is equivalent to the following LP problem:
\begin{align}
\label{eq:map_marginal_poly}
\min_{ \{\by,\bY \} \in \mathcal{M}(\mathcal{G},\setZ) } \mathrm{E}(\by,\bY) &:=
\sum_{p\in \setV, i \in \setZ}
 \theta_p(i) y_{p,i} \,\, + \\
&\sum_{p < q, (p,q) \in \setE, i,j \in \setZ}
\theta_{p,q}(i,j) Y_{pi,qj} \notag
\end{align}
where the set $\mathcal{M}(\mathcal{G,\setZ})$ refers to the {\em marginal polytope} \wrt graph $\mathcal{G}$ and the state space $\setZ$:
\begin{align}
 &\mathcal{M}(\mathcal{G}, \setZ) :=  \\
 &\,\, \left\{ \by, \bY \
              \begin{array}{|lll}
                 & \exists \,  \mathrm{Pr}(\bx), \, \sst    &   \\
                 & y_{p,i}   \!=\! \mathrm{Pr}(x_p \!=\! i),          & \forall     p \!\in\! \setV, i \!\in\! \setZ, \\
                 & Y_{pi,qj} \!=\! \mathrm{Pr}(x_p \!=\! i, x_q \!=\! j), & \forall (p,q) \!\in\! \setE, i \!\in\! \setZ.\!
              \end{array}
      \right\} \notag
\end{align}
In the above definition, $\{ y_{p,i} \}_{ i \in \setZ}$ and $\{ Y_{pi,qj} \}_{i,j \in \setZ }$
refer to singleton and pairwise pseudomarginals \wrt each node $p
\!\in\! \setV$ and each edge $(p,q) \!\in\! \setE$ respectively,
which are globally realizable by some joint distribution $\mathrm{Pr}()$ over
$\bx \in \setZ^n$.
Each vertex (extreme point) of $\mathcal{M}(\mathcal{G},\setZ)$ is integer
($\{0,1 \}$-valued)
and corresponds to a valid assignment $\bx \in \setZ^n$.
So there is always a binary optimal solution to the LP problem \eqref{eq:map_marginal_poly}.

Although the above LP problem only contains $\mathcal{O}(|\mathcal{E}|+ |\mathcal{V}|)$ number of variables,
it generally requires exponential number of linear constraints to describe $\mathcal{M}(\mathcal{G},\setZ)$.
To address this difficulty,
the standard LP relaxation\ \cite{shlezinger1976syntactic,werner2007linear,wainwright2008graphical}
optimizes the objective function of \eqref{eq:map_marginal_poly} over a tractable set, called {\em local marginal polytope}:
\begin{align}
 &\mathcal{M}_L(\mathcal{G},\setZ) \!:=\! \label{eq:lmp}\\
 &\,\, \left\{ \by, \bY \
              \begin{array}{|lll}
                 &\ \by \geq \mathbf{0}; \ \bY \geq \mathbf{0}; & \\
                 &\ {\sum_{i\in\setZ}} y_{p,i} \!=\! 1,         & \forall p \!\in\! \setV;   \\
                 &\ {\sum_{j\in\setZ}} Y_{pi,qj} \!=\! y_{p,i}, & \forall (p,q) \!\in\! \setE, i \!\in\! \setZ.
              \end{array}
      \right\}, \notag
\end{align}
which
is a convex polyhedral outer-bound on
$\mathcal{M}(\mathcal{G},\setZ)$.
That is, $\mathcal{M}(\mathcal{G},\setZ) \subseteq
\mathcal{M}_L(\mathcal{G},\setZ)$.
Its number of linear constraints
grows linearly \wrt the graph size but quadratically with the size of $\setZ$.
It is known (see \cite{wainwright2008graphical} for example) that
any vertex of $\mathcal{M}(\mathcal{G},\setZ)$ is also a vertex of $\mathcal{M}_L(\mathcal{G},\setZ)$,
and for a graph $\mathcal{G}$ with cycles, $\mathcal{M}_L(\mathcal{G},\setZ)$ also contains
fractional vertexes lying outside $\mathcal{M}(\mathcal{G},\setZ)$.
Consequently, we have the following important property of the standard LP relaxation
(see \cite{wainwright2008graphical}):
\begin{theorem}
\label{thm:lp}
{(Optimality guarantee of LP relaxation)}

If the solution to the LP relaxation: $\displaystyle{\min_{\{\by,\bY \} \in \mathcal{M}_L(\mathcal{G},\setZ)}} \mathrm{E}(\by,\bY)$ is integer,
it is the exact MAP solution.
\end{theorem}

The edge consistency is preserved in
$\mathcal{M}_L(\mathcal{G},\setZ)$,
but high-order consistency is not enforced.
Cluster based methods~\cite{sontag2008,batra2011tighter,sontag2012efficiently} iteratively add to LP relaxation
high-order consistency constraints,
and provide incrementally improv\-ed lower-bounds to the minimal energy:
$\min_{\bx \in \setZ^n} \mathrm{E}(\bx)$.

Alternatively, SDP relaxation provides
a different convex outer-bound on  $\mathcal{M}(\mathcal{G},\setZ)$.
Consider a symmetric matrix consisting of $\by$ and $\bY$:
\begin{align}
\label{eq:yY}
\Omega(\by,\bY) \!:=\!
\left[{ \begin{smallmatrix}%
  1       &y_{1,1}    &\cdots &y_{p,i}   &\cdots &y_{q,j}   &\cdots &y_{n,h} \\
  y_{1,1} &Y_{11,11}  &\cdots &Y_{11,pi} &\cdots &Y_{11,qj} &\cdots &Y_{11,nh} \\
  \tvdots  &\tvdots     &       &\tvdots    &       &\tvdots    &       &\tvdots \\
  y_{pi}  &Y_{pi,11}  &\cdots &Y_{pi,pi} &\cdots &Y_{pi,qj} &\cdots &Y_{pi,nh} \\
  \tvdots  &\tvdots     &       &\tvdots    &       &\tvdots    &       &\tvdots \\
  y_{qj}  &Y_{qj,11}  &\cdots &Y_{qj,pi} &\cdots &Y_{qj,qj} &\cdots &Y_{qj,nh} \\
  \tvdots  &\tvdots     &       &\tvdots    &       &\tvdots    &       &\tvdots \\
  y_{nh}  &Y_{nh,11}  &\cdots &Y_{nh,pi} &\cdots &Y_{nh,qj} &\cdots &Y_{nh,nh}
\end{smallmatrix}}\right],
\end{align}
where $Y_{pi,qj} := y_{p,i} \!\cdot\! y_{q,j}, \forall (p,q) \!\notin\! \setE, i,j \!\in\! \setZ$.
At any vertex of $\mathcal{M}(\mathcal{G},\setZ)$,
$\Omega(\by,\bY) = \Omega(\by,\by \by^\T) =
\left[ \begin{smallmatrix} 1,\hfill& \by^\T \hfill\\ \by,\hfill& \by\by^\T \end{smallmatrix} \right]$ $ \psdd \mathbf{0}$
is a rank-$1$ positive semidefinite matrix.

By discarding the non-convex rank-$1$ constraint,
we obtain the following convex outer-bound on $\mathcal{M}(\mathcal{G},\setZ)$:
\begin{align}
&\mathcal{P}(\mathcal{G}) := \{ \by, \bY \lvert \Omega(\by,\bY) \psdd \mathbf{0}\},
\end{align}
which generally contains curved boundaries.
It is shown in \cite{wainwright2008graphical} that local marginal polytope $\mathcal{M}_L(\mathcal{G},\setZ)$
and the semi\-definite outer-bound $\mathcal{P}(\mathcal{G})$
are mutually incomparable,
whi\-ch means that neither of them dominates the other.
{
  We attempt to combine the local marginal polytope and semidefinite outer-bound
in this work such that a tighter bound can be achieved.
To solve the resulted SDP problem, which may have a large number of constraints,
 scalable SDP solvers are developed.

In general, both LP and SDP relaxation do not provide feasible integer solutions to the original
non-submodular energy minimization problem.
Thus a rounding procedure is needed to transform a fractional solution given by LP/SDP relaxation to a feasible integer solution.
A simple rounding procedure for both LP/SDP relaxation is to round
each variable independently based on the singleton pseudo-marginals $\by$:
\begin{align}
x_p = \arg \max_{i \in \setZ} y_{p,i}, \,\, p = 1,2,\cdots,n. \label{eq:simple_rounding}
\end{align}
Better rounding methods would consider the labelling over all variables jointly,
which can be done by considering the reparameterized objective~\cite{kolmogorov2006convergent} or
pairwise pseudo-marginals $\bY$ conditioned by already labelled variables.
For max-cut problems, the random hyperplane rounding approach~\cite{Goemans95improved} for SDP relaxation
delivers a bounded expected objective value.
In this paper, the simple rounding procedure \eqref{eq:simple_rounding}
is adopted for our SDP solver,
followed by a post-processing procedure based on ICM~\cite{besag1986statistical} (see Section~\ref{sec:sdp_speedup}).}

\section{Branch-and-Cut for Energy Minimization}
\label{sec:bcsdp}

{

In this section, the overall algorithm of \BC for the energy minimization problem \eqref{eq:energy0} is
given in Algorithm~\ref{alg:bb} and then the subproblem selection and branching strategies are presented.

\subsection{The Branch-and-Bound Method}
First, we briefly revisit \BB methods~\cite{horst2000introduction,hendrix2010introduction}.
As a method for  global optimization of nonconvex problems,
\BB methods rely on two procedures
which compute
upper- and lower-bounds on the global optimum.
Considering a minimization problem,
the upper-bound can be obtained by choosing any point in the feasible set.
A local search method is usually performed to improve the upper-bound.
On the other hand, the lower-bound can be computed from a convex relaxation to the original non-convex problem.

At each iteration of \BB (Algorithm~\ref{alg:bb}), a subproblem is selected from the priority queue $\setQ$ (Step $1$)
and its upper-/lower-bounds are computed (Step $2$).
In Step $3$, the global upper-bound ($\mathrm{gub}$) is updated in each iteration as the minimum of the upper-bounds over all branches,
while the global lower-bound ($\mathrm{glb}$) is updated as the minimum of the lower-bounds over all leaf branches.
All subproblems with a lower-bound not smaller than the global upper-bound are pruned in Step $4$.
If the selected subproblem cannot be pruned, its feasible set $\setD$
is separated into at least two convex sets (Step $5$).
\BB algorithms terminate when the global upper-bound and lower-bound are sufficiently close to each other.

There are several essential components in \BB:
bounding ($\mathrm{bound}(\setD)$),
subproblem selection ($\mathrm{pop}(\setQ)$)
and branching ($\mathrm{split}(\setD,\by)$).
The bounds computed by the bounding procedure are assumed to become tighter as the feasible set shrink.
Especially when the size of a feasible set shrinks to a point, the difference
between upper and low er bounds needs to converge to zero.
It is shown in Proposition~\ref{prop1} that our SDP bounding approach satisfies these assumptions.
A comprehensive study of branching and subproblem selection strategies can be found in
\cite{achterberg2005branching,Computercodes1979,linderoth1999computational,mitra1973investigation},
in which a number of sophisticated rules are investigated.
In this work,
we  adopt simple rules for branching and subproblem selection
and rely on the tight bounding procedure to improve the performance.

   \begin{algorithm}[t]
   \footnotesize
   \setcounter{AlgoLine}{0}
   \caption{Semidefinite Branch-and-Cut (SDBC) }%
   \centering
   \begin{minipage}[]{1\linewidth}
   \KwIn{ the original energy minimization
        problem $\min_{\bx \in \setZ^n} \fE(\bx)$.
    }
       { {\bf Initialization}:
       global upper-/lower-bounds $\GUB = +\infty$,
       $\GLB = -\infty$;
       priority queue $\setQ = \{ \setZ^n \}$.

   }

   \While{ $\setQ \neq \emptyset$ }
   {
     {\bf Step 1 }%
     \hangindent 1.3em
     {\em Subproblem Selection}: $ \setD = \mathrm{pop}(\setQ)$.

     {\bf Step 2 } %
     \hangindent 1.3em
     {\em Bounding}: \\
     - $[\UB, \LB, \bx, \by]
        = \mbox{bound}(\setD)$
       (see Algorithm.~\ref{alg:bound});

     {\bf Step 3 }%
     \hangindent 1.3em
     { Update
     global bounds}:\\
     - $\GLB = \min(\LB, \min_{\setD' \in \setQ} \LB(\setD'))$;\\
     - \lIf{$\GUB > \UB$}
       {$\GUB = \UB$, $\bx^{\star} = \bx$.}

     {\bf Step 4 }%
     \hangindent 1.3em
     {\em Pruning}: $\setQ \leftarrow \setQ \backslash
     \{\setD' | \setD' \in \setQ,
     \LB(\setD') \geq \GUB \}$.

     {\bf Step 5 }%
     \hangindent 1.3em
     {\em Branching}: \\
                         \If( ){ $\LB < \GUB$}
     { - $[\setD_{1}, \setD_{2}] = \fbranch (\setD,\by)$
         (see Algorithm~\ref{alg:branch}); \\
       - $\setQ \leftarrow \{ \setQ, \setD_{1}, \setD_{2}\}$.}

   }
   \KwOut{$\bx^{\star}$.
   }
   \end{minipage}
   \vspace{-0.1cm}
   \label{alg:bb}
   \end{algorithm}

   \begin{algorithm}[t]
   \footnotesize
   \setcounter{AlgoLine}{0}
   \caption{The branching procedure,
            $[\setD_1,\setD_2] = \mathrm{split}(\setD,\by)$.}
   \centering
   \begin{minipage}[]{1\linewidth}
   {
   \KwIn{The search space $\setD := \setZ^{\setD}_1 \times \setZ^{\setD}_2 \times \dots \times \setZ^{\setD}_{|\setV|}$,
       and the singleton pseudomarginals $\{y_{p,i}\}_{i \in \setZ^{\setD}_{p}}, \forall p \in \setV $.}

   {\bf Step 1 }%
   \hangindent 1.3em
   {\em Node selection}:
   $\varphi = \arg \displaystyle{\min_{p \in \setV, |\setZ^{\setD}_p| > 1}} ( \max_{i\in \setZ^{\setD}_{p}} y_{p,i} )$.

   {\bf Step 2 }%
   \hangindent 1.3em
   {\em State ordering}: obtain the state order $i_1, i_2, \dots, i_{|\setZ^{\setD}_{\varphi}|}$ such that
                         $y_{\varphi,i_1} \geq y_{\varphi,i_2} \geq \dots \geq y_{\varphi,i_{|\setZ^{\setD}_{\varphi}|}}$.

   {\bf Step 3 }%
   \hangindent 1.3em
   {\em Branching}:
   separate $\setZ^{\setD}_{\varphi}$ into two disjointed subsets
$\setZ^{\setD}_{\varphi,1} = \{i_1, \dots,  i_{\lfloor {|\setZ^{\setD}_{\varphi}|}/{2} \rfloor} \}$ and
$\setZ^{\setD}_{\varphi,2} = \{i_{\lfloor {|\setZ^{\setD}_{\varphi}|}/{2} \rfloor+1}, \dots i_{|\setZ^{\setD}_{\varphi}|} \}$.

   \KwOut{ $\setD_1 \!=\! \setZ^{\setD}_{1} \times \dots \times \setZ^{\setD}_{\varphi,1} \times \dots \times \setZ^{\setD}_{|\setV|}$ and
           $\setD_2 \!=\! \setZ^{\setD}_{1} \times \dots \times \setZ^{\setD}_{\varphi,2} \times \dots \times \setZ^{\setD}_{|\setV|}$.

   }
   }
   \end{minipage}
   \label{alg:branch}
   \end{algorithm}

\subsection{Subproblem Selection: $\setD = \mathrm{pop}(\setQ)$}

All  unresolved subproblems (represented by their respective feasible sets) are stor\-ed in the priority queue $\setQ$.
At each iteration of \BB, the subproblem with the highest priority will be selected and split.
In our implementation, subproblems are sorted according to the quality of their lower-bounds,
and the subproblem with the worst lower-bound will be split firstly.
{
In other words,
\begin{align}
\setD = \mathrm{pop}(\setQ) = \arg \min_{\setD' \in \setQ} \mathrm{lb} (\setD'),
\end{align}
where $\setD := \setZ^{\setD}_1 \times \setZ^{\setD}_2 \times \dots \times \setZ^{\setD}_{|\setV|}$ and
$\setZ^{\setD}_{p}$ denotes the set of possible states for each node $p \in \setV$ within the feasible set $\setD$.
Note that a subset of elements of $\by$ are enforced to be $0$s or $1$s in the subproblem with respect to $\setD$:
\begin{subequations}
\label{eq:reparameterization}
\begin{align}
y_{p,i} = 0, &\quad \mbox{if } i \in \{\setZ \backslash \setZ^{\setD}_p \}, \forall p \in \setV, \\
y_{p,i} = 1, &\quad \mbox{if } i \in \setZ^{\setD}_p \mbox{ and } |\setZ^{\setD}_p| = 1, \forall p \in \setV.
\end{align}
\end{subequations}
}

\subsection{Branching: $[\setD_1, \setD_2 ] = \mathrm{split}(\setD,\by)$}
\label{sec:branch}

{ At the branching step, the input feasible set $\setD$
is further split into two disjointed sets $\setD_1$ and $\setD_2$, by selecting a node and split its state space in halves.}

\noindent
{\bf Node Selection}
We employ the ``difficult first" strategy, which
selects the node that is the most difficult to discretize.
Remind that $\{y_{p,i}\}_{i \in \setZ^{\setD}_{p}}$ can be considered as a singleton pseudo-marginal for each node $p \in \setV$,
if it resides in the local marginal polytope \eqref{eq:lmp}.
If this distribution does not concentrate on a single point (in other words, spreads evenly over all states),
then the corresponding node is considered to be difficult to discretize.
{
Accordingly, a simple strategy is adopted for node selection in this work:
\begin{align}
\varphi = \arg \displaystyle{\min_{p \in \setV, |\setZ^{\setD}_p| > 1}} ( \max_{i\in \setZ^{\setD}_{p}} y_{p,i} ).
\end{align}
}

\noindent
{\bf States Ordering}
{ As in \cite{Sun2012MRFBB2},
the possible states of the selected node $\varphi$, \ie $i_1, i_2, \dots, i_{|\setZ^{\setD}_{\varphi}|}$,
are ordered dynamically in each iteration of branching,
such that $y_{\varphi,i_1} \geq y_{\varphi,i_2} \geq \dots \geq y_{\varphi,i_{|\setZ^{\setD}_{\varphi}|}}$.
Then the most and the least probable states are separated into two sets:
$\setZ^{\setD}_{\varphi,1} = \{i_1, \dots,  i_{\lfloor {|\setZ^{\setD}_{\varphi}|}/{2} \rfloor} \}$ and
$\setZ^{\setD}_{\varphi,2} = \{i_{\lfloor {|\setZ^{\setD}_{\varphi}|}/{2} \rfloor+1}, \dots i_{|\setZ^{\setD}_{\varphi}|} \}$.}

Now the last component in the \BC approach (Algorithm~\ref{alg:bb})
to be studied is the bounding procedure, $\mathrm{bound}(\setD)$,
which we discuss in the following section.

} %

\section{An Efficient SDP Bounding Procedure with Cutting-plane}
\label{sec:sdp_relaxation}

{
In this section, we consider the following SDP relaxation to the original energy minimization problem~\eqref{eq:energy0}:
\begin{subequations}
\label{eq:map}
\begin{align}
\min_{\by,\bY} \,\, & \mathrm{E}(\by,\bY) \\ %
\sst \,\,
&
\eqref{eq:sdprelax_cons1},
\eqref{eq:sdprelax_cons2},
\eqref{eq:sdprelax_cons3},
\eqref{eq:sdprelax_cons4},
\eqref{eq:sdprelax_cons5},
\eqref{eq:tricons}, \eqref{eq:cyccons}, \eqref{eq:owccons}, \label{eq:map_cons} \\
& \Omega(\by, \bY) \psdd \mathbf{0},
\end{align}
\end{subequations}
where \eqref{eq:map_cons} corresponds to the additional linear constraints to be discussed in Section~\ref{eq:addi_cons}.}
The above formulation can be easily extended to any subproblem $\setD$ in Algorithm~\ref{alg:bb},
by reparameterization or enforcing additional constraints \eqref{eq:reparameterization}.
A main observation is that the linear constraints arising from standard/high-order LP relaxation
can be used to significantly tighten the following naive SDP relaxation:
However, the resulted large-scale SDP optimization problem,
which in particular has a large number of linear constraints, render standard interior-point SDP solvers
inapplicable.
A scalable SDP solver combined with cutting-plane is proposed to solve such SDP optimization in Section~\ref{sec:initbounding} and \ref{eq:cutting-plane}.
Then the complexity of our method is discussed in Section~\ref{sec:complexity} and several speeding-up strategies are proposed in Section~\ref{sec:sdp_speedup}.

\subsection{Linear Constraints that Tighten the Semidefinite Outer-Bound}
\label{eq:addi_cons}
Besides the \psd constraint, several classes of linear constraints arising from LP relaxation
can be added to SDP relaxation to tighten
the bound.

\noindent
{\bf $\mathbf{0\slash1}$-Integer Constraints} These constraints arise from
the integer constraints $y_{p,i} \!\in\! \{ 0,1 \}$,
which is equivalent to $y_{p,i} \!=\! y_{p,i}^2$, for each node $p \in \setV$ and state $i \in \setZ$:
\begin{align}
\label{eq:sdprelax_cons1}
Y_{pi,pi} = y_{p,i}, \,\,\forall p \in \setV, i \in \setZ.
\end{align}
There are $nh$ such constraints for a graph with $n$ nodes and $h$ states per node.

\noindent
{\bf Local Normalization Constraints} Because only one state
$i \in \setZ$ can be assigned to each node $p \in \setV$,
we have:
\begin{align}
\label{eq:sdprelax_cons2}
\textstyle{\sum_{i \in \setZ}} y_{p,i} = 1.
\,\,\forall p \in\setV,
\end{align}
There are $n$ local normalization constraints in total.
Note that
for any $\{\by,\bY \} \in \mathcal{P}(\mathcal{G}) \cap \eqref{eq:sdprelax_cons1} \cap \eqref{eq:sdprelax_cons2}$,
if $\mathrm{rank}(\Omega(\by,\bY)) = 1$, then $\{ \by,\bY \}$ is a vertex of $\mathcal{M}(\mathcal{G},\setZ)$.

\noindent
{\bf Non-Negativity Constraints}
Deriving from $Y_{pi,qi} = y_{p,i} \cdot y_{q,j}$ and $\by \in \{0,1 \}^{nh}$, the following constraints holds:
\begin{align}
Y_{pi,qj} \geq 0, \,\,\forall (p,q) \in \setE, i,j \in \setZ.
\label{eq:sdprelax_cons3}
\end{align}
There are $\lvert \mathcal{E} \rvert h^2$ such constraints for a graph with $\lvert \mathcal{E} \rvert$ edges.
The number becomes $n(n-1)h^2/2$ for a fully\--connected graph.
Huang \etal~\cite{guibasscalable} have shown that without the non-negativity constraints,
SDP relaxation is loose for submodular functions.

\noindent
{\bf Edge Marginalization Constraints}
For each edge $(p,q) \in \setE$ and $i \in \setZ$,
we have $\sum_{j \in \setZ} Y_{pi,qj} = y_{pi} \cdot \sum_{j \in \setZ} y_{qj}$.
Therefore the following constraints can be derived:
\begin{align}
  \!{\textstyle \sum_{j\in\setZ} Y_{pi,qj} = y_{p,i}, \,\, \forall (p,q) \in \setE, i \in \setZ.}
\label{eq:sdprelax_cons4}
\end{align}
There are in total $2 \lvert \mathcal{E} \rvert h$ such constraints ($n(n-1)h$ for fully-connected graphs).
Note that local normalization~\eqref{eq:sdprelax_cons2},
non-negativity~\eqref{eq:sdprelax_cons3}
and marginalization~\eqref{eq:sdprelax_cons4}
constraints force $\{ \by, \bY \}$ to lie in local marginal polytope $\mathcal{M}_L(\mathcal{G},\setZ)$.

\noindent
{\bf Gangster Operators}
Because $Y_{pi,pj} = y_{p,i} \cdot y_{p,j}$ and each node $p \in \setV$ can only be
assigned with one state $i \in \setZ$,
at least one of $y_{p,i}$ and $y_{p,j}$ should be zero.
Then we have the following constraints:
\begin{align}
Y_{pi,pj} = 0,  \,\, \forall  p \in \setV, \forall  i \neq j, i,j \in \setZ. \label{eq:sdprelax_cons5}
\end{align}
These constraints are referred to as {\em gangster operators}, as they shoot holes (zeros) in $\bY$.
Considering matrix symmetry, there are $nh(h-1)/2$ such constraints in total.
Schellewald and Schnorr~\cite{Schellewald05} have used these constraints for SDP approaches to subgraph matching.

Note that the above linear constraints are directly defined on the marginal polytope. In this work,
we also consider the following a few classes of linear constraints defined on the cut polytope
or equivalently binary marginal polytope, which can be extended to $\mathcal{M}(\setG,\setZ)$ by projecting a non-binary
graph to a binary graph.
Comprehensive studies of linear constraints for cut polytope or binary marginal polytope can be found
in \cite{barahona1986cut,deza1992clique,chopra1993partition} and the references therein.

{

\noindent
{\bf Triangular Inequalities}
Considering three binary variables $\pi_p$, $\pi_q$, $\pi_r$ and
define $\delta_{p,q} := \mathbf{1}[\pi_p \neq \pi_q]$, then we have:
\begin{subequations}
\label{eq:tricons}
\begin{align}
\delta_{p,q} + \delta_{q,r} + \delta_{p,r} \leq 2, \\
\delta_{p,q} - \delta_{q,r} - \delta_{p,r} \leq 0, \label{eq:tricons_2}\\
- \delta_{p,q} + \delta_{q,r} - \delta_{p,r} \leq 0, \label{eq:tricons_3}\\
- \delta_{p,q} - \delta_{q,r} + \delta_{p,r} \leq 0, \label{eq:tricons_4}
\end{align}
\end{subequations}
which are facet defining for a cut polytope~\cite{barahona1986cut}.
The number of triangular inequalities is cubic in the number of binary variables.

\noindent
{\bf Cycle Inequalities} %
They
arise from the fact
that there must be an even number of edge-cuts for any cycle of variables.
For a cycle $C$ and any $F \subseteq C$ such that $|F|$ is odd, the inequality can be expressed as:
\begin{align}
\label{eq:cyccons}
\sum_{(p,q) \in C \backslash F} \delta_{p,q} + \sum_{(p,q) \in F} (1-\delta_{p,q}) \geq 1.
\end{align}
The triangular inequalities \eqref{eq:tricons_2} \eqref{eq:tricons_3}, \eqref{eq:tricons_4}
can be considered as special cases of cycle inequalities.
A cycle inequality is  facet defining for a cut polytope if and only if the cycle is chordless~\cite{barahona1986cut}.
There are an exponential number of cycle inequalities in a graph.
However, the most violated one can be found in $\mathcal{O}(n^3h^3)$ time for a dense graph
using the separation procedure shown in \cite{barahona1986cut,deza1996geometry}.

\noindent
{\bf Odd-wheel Inequalities}
Consider a cycle $C$ of length $s=2t+3$, where the integer $t \geq 0$.
$V(C)$ denotes the node set of $C$ and $\pi_r$ refers to a center node not in $V(C)$.
Then the following inequality
\begin{align}
\label{eq:owccons}
\sum_{(p,q) \in C} \delta_{p,q} - \sum_{p \in V(C)} \delta_{p,r} \leq \frac{|C|-1}{2},
\end{align}
is called `$s$-wheel inequality' or simply `odd-wheel inequality'.
This class of constraints is not facet-defining for a cut polytope,
but it is facet defining for a multicut polytope~\cite{deza1992clique,chopra1993partition}.
There are also an exponential number of the above constraints with respect to the number of binary variables.
For a solution satisfying all cycle inequalities, the separation procedure shown in \cite{deza1996geometry}
can find the most violated odd-wheel inequality using $\mathcal{O}(n^4h^4)$ time for a dense graph.

We adopt the method proposed in \cite{sontag2007new,sontag2012efficiently} for
projecting non-binary graphs to binary graphs.
Firstly, $\pi_p^s$ is defined as a binary variable separating the state space of node $p$
into two disjoint non-empty sets $\setZ_s$ and $\setZ\! \setminus \! \setZ_s$,
that is $\pi_p^s = 1$, if $x_p \in \setZ_s$,  and $0$ otherwise.
$\pi_p = \{ \pi_p^1, \pi_p^2, \dots \}$ represents a collection of partitions \wrt node $p$.
Then the non-binary graph $\setG = (\setV, \setE)$ can be projected to a binary graph $\setG_\pi = (\setV_\pi, \setE_\pi)$, where
\begin{subequations}
\label{eq:prj_binary}
\begin{align}
\setV_{\pi} &= \{\pi_p^s \mid p \in \setV, s \leq |\pi_p| \}, \\
\setE_{\pi} &= \{ (\pi_q^s, \pi_p^t) \mid (p,q) \in \setE,  s \leq |\pi_p| , t \leq |\pi_q| \}.
\end{align}
\end{subequations}
Note that $\delta_{\pi_q^s,\pi_p^t} := \mathbf{1}[ \pi_p^s \neq \pi_q^t]$ can be expressed
in terms of $\bY$ in \eqref{eq:yY},
that is $\delta_{\pi_q^s,\pi_p^t} =
\sum_{ \{i \in \setZ_s, j \notin \setZ_t \} \cap \{i \notin \setZ_s, j \in \setZ_t \} } Y_{pi,qj}$.
In this paper, we define $\setZ_s = \{ s \}, \forall s = 1, 2, \cdots, h$, such that
each $h$-state variable $x_p$ is projected to $h$ binary variables.

Note that the projected inequalities are not necessarily facet-defining for the non-binary graph,
although some of them are facet-defining on the projected binary graph.

}

   \begin{algorithm}[t]
   \footnotesize
   \setcounter{AlgoLine}{0}
   \caption{SDP Bounding Procedure
           ($[\mathrm{ub}, \mathrm{lb},
            \bx, \by] = \mathrm{bound}(\setD)$)}

   \centering
   \begin{minipage}[]{1\linewidth}
{
   \KwIn{ The energy minimization problem
          $\min_{\bx \in \setD} \mathrm{E}(\bx)$, where $\setD$ is a subset of the space of potential label assignment $\setZ^n$;
          $\Psi$ is the working set of constraints.
          $\Psi_{\mathrm{addi}}$ is the pool of additional constraints
          \eqref{eq:sdprelax_cons3},
          \eqref{eq:sdprelax_cons4},
          \eqref{eq:tricons}, \eqref{eq:cyccons} and \eqref{eq:owccons}
          to be added in cutting-plane.
          Parameters $\gamma \!>\! 0$, $\delta \!>\! 1$;
          Maximum iterations
          $K_{\mathrm{init}}$,
          $K_{\mathrm{inner}}$ and $K_{\mathrm{outer}}$.
          The initial set of constraints $\psi_{\mathrm{init}}$ and the corresponding dual variable $\bu_{\mathrm{init}}$.}

   {\bf Step 1 } {\em Initial Bounding}: \\
   - Model reduction (Optional): $\setD \leftarrow \mathrm{QPBO}(\setD)$; \\
   - Initialize the working set and dual variable: $\Psi \leftarrow \psi_{\mathrm{init}}$, $\bu \leftarrow \bu_{\mathrm{init}}$. \\
   - Construct the dual problem \eqref{eq:fastsdp_dual}
     \wrt $\setD$, $\Psi$ and $\gamma$. \\
   \For{$k = 1, 2, \dots, K_{\mathrm{init}}$}
   {
     - Dual variable update:
       $\bu \leftarrow \bu +
     \rho \bH \nabla \mathrm{d}_{\gamma}(\bu)$.   \\
     - Lower-bound update:
     $\mathrm{lb} \leftarrow \mathrm{d}_\gamma (\bu)$.
     \\ - Goto {\bf Step 3} if any stop condition is met (see Section~\ref{sec:sdp_speedup}).
   }
   - Compute the primal variable: $\Omega(\by, \bY)
      = \gamma \Pi_{\mathcal{S}^{n\!h\!+\!1}_+} (\bC(\bu))$.

   {\bf Step 2} {\em Cutting-Plane}: \\
   \For{$k_1 = 1, 2, \dots, K_{\mathrm{outer}}$}
   {
     - $\psi_{\mathrm{in}} \leftarrow$ inactive constraints selected from $\Psi$. \\
     - $\psi_{\mathrm{vi}} \leftarrow$ constraints violated by $\Omega(\by,\bY)$ selected from $\Psi_{\mathrm{addi}}$. \\
     - $\Psi \leftarrow \{ \Psi \backslash \psi_{\mathrm{in}} \} \cup \psi_{\mathrm{vi}}  $.
       $\gamma \leftarrow \delta \cdot \gamma$. \\
     - Construct the dual problem~\eqref{eq:fastsdp_dual}.
       \wrt $\setD$, $\Psi$ and $\gamma$.\\
     - Update the dual variable:
       $\bu \leftarrow \{\bu, \mathbf{0}\}$, where
       $\mathbf{0}$ corresponds to the dual variables
       \wrt $\psi$. \\
     \For{$k_2 = 1, 2, \dots, K_{\mathrm{inner}}$}
     {
     - Dual variable update:
       $\bu \leftarrow \bu +
     \rho \bH \nabla \mathrm{d}_{\gamma}(\bu)$. \\
     - Lower-bound update:
     $\mathrm{lb} \leftarrow \mathrm{d}_\gamma (\bu)$.
     }
     - Goto {\bf Step 3} if any stop condition is met. \\
     - Compute the primal variable: $\Omega(\by, \bY)
      = \gamma \Pi_{\mathcal{S}^{n\!h\!+\!1}_+} (\bC(\bu))$.
   }

   {\bf Step 3} {\em Rounding}: \\
   - Generate a discrete solution $\bx$
   by rounding $\by$ (see Equation~\eqref{eq:simple_rounding}). \\ %
   - Improve $\bx$ using RICM (see Section~\ref{sec:ricm}). \\
   - $\mathrm{ub} \leftarrow \fE(\bx)$.

   \KwOut{ $\mathrm{ub}$, $\mathrm{lb}$,
           $\bx$ and $\by$.
   }
}

   \end{minipage}
   \label{alg:bound}
   \end{algorithm}

\subsection{Initial Bounding Using SDCut}
\label{sec:initbounding}

The feasible set of the SDP relaxation~\eqref{eq:map} is the intersection of
semidefinite outer-bound $\mathcal{P}(\setG)$ and the polyhedral
outer-bound produced by standard/high-order LP relaxation.
There are several difficulties in optimizing the SDP problem~\eqref{eq:map}.
First of all, it is computationally inefficient to solve
SDP problems using classic interior-point methods,
which has the computational complexity of $\mathcal{O}(m^3+mn^3h^3+m^2n^2h^2)$
and memory requirement of $\mathcal{O}(m^2+n^2h^2)$, where $m$ is the number of linear constraints.
Secondly, the linear constraints involved in \eqref{eq:map} are  very complex.
It requires $\mathcal{O}(n^2h^2)$ constraints to
describe the local marginal polytope for a fully-connected graph,
which makes the computational complexity of interior-point methods increase to $\mathcal{O}(n^6h^6)$.
The total number of cycle inequalities grows even exponentially with the graph size.
To this end, we propose to adapt the SDCut method proposed in our prior work \cite{peng2013cpvr}
to optimize \eqref{eq:map} and use cutting-plane to cope with the large number of linear constraints.

In the following, $\bA, \bB_i \in
\mathcal{S}^{nh+1}, i = 1,2,\cdots,m$, $\bb \in \mathbb{R}^m$ are
defined such that $\langle \Omega(\by,\bY), \bA \rangle = \mathrm{E}(\by,\bY)$
and linear constraints in \eqref{eq:map_cons} are also expressed in the
form of matrix inner-products:
$ \langle  \Omega(\by,\bY), \bB_i   \rangle =    b_i, \,\forall i \in \mathcal{I}_{eq}$,
$ \langle  \Omega(\by,\bY), \bB_i   \rangle \leq b_i, \,\forall i \in
\mathcal{I}_{in}$,
where $\mathcal{I}_{eq}$, $\mathcal{I}_{in}$ refer to
non-overlapping indexes for linear equality and inequality constraints
respectively.
$m \!=\! |\mathcal{I}_{eq}| + |\mathcal{I}_{in}|$.

SDCut~\cite{peng2013cpvr} is proposed to {approximately} solve a subclass of SDP problems
in which the trace of \psd matrix variables is fixed.
Instead of minimizing the linear objective function
$\langle \Omega(\by,\bY), \bA \rangle$,
SDCut minimizes the perturbed function  $\langle \Omega(\by,\bY), \bA \rangle +
\frac{1}{2\gamma}(\lVert \Omega(\by,\bY) \rVert^2_F - \eta^2)$
over the constraints shown in \eqref{eq:map_cons},
where $\eta := \mathrm{trace}(\Omega(\by,\bY)) = n\!+\!1$ ($ \eta  $ can be discarded since it is a constant)
in our case and $\gamma > 0$.
It is shown in \cite{peng2013cpvr} that with an appropriate $\gamma$,
the solution of the perturbed problem is ``close'' to that of the
original SDP problem.

Furthermore, by introducing the perturbed objective,
the Lagrangian dual of SDCut can be simplified to:
\begin{subequations}
\label{eq:fastsdp_dual}
\begin{align}
\max_{\bu \in \mathbb{R}^m} \,\,
     &\mathrm{d}_\gamma(\bu) \!:=\! - \frac{\gamma}{2} \lVert \Pi_{\mathcal{S}^{n\!h\!+\!1}_+} (\bC(\bu)) \rVert_F^2
     \!-\! \bu^{\T} \bb \!-\! \frac{\eta^2}{2\gamma}, \label{eq:dual_obj}\\
\sst       \,\, &u_i \geq 0, i \in \mathcal{I}_{in},
\end{align}
\end{subequations}
where $\bC(\bu) \!=\! - \bA - \sum_{i=1}^{m} u_i \bB_i$.
The matrix-valued function $\Pi_{\mathcal{S}^{n\!h\!+\!1}_+}$ is defined as
\[
  \Pi_{\mathcal{S}^{n\!h\!+\!1}_+}
(\bX) := \bP \mathrm{diag}(\max(0,\mathbf{\boldsymbol \lambda})) \bP^\T,
\]
and $\bX = \bP \mathrm{diag}({\boldsymbol \lambda}) \bP^\T$ is the eigen-decomposition of the matrix $\bX \in \mathcal{S}^n$,
where ${\boldsymbol \lambda}$ denotes the vector of eigenvalues and $\bP$ contains the corresponding (column) eigenvectors.
Note that there is no \psd constraint in \eqref{eq:fastsdp_dual} any more.
Based on K.K.T.\ conditions, we have
\begin{align}
\Omega(\by^{\star}_\gamma, \bY^{\star}_\gamma)
 = \gamma \Pi_{\mathcal{S}^{n\!h\!+\!1}_+} (\bC(\bu_\gamma^\star)), \label{eq:fastsdp_primal_dual}
\end{align}
where $\by_\gamma^\star$,$\bY_\gamma^\star$ and $\bu_\gamma^\star$ are the optimal solutions to the corresponding primal and dual problems \wrt $\gamma$ respectively.

We demonstrate in Appendix~\ref{app:1} that the dual formulation of SDCut~\eqref{eq:fastsdp_dual}
is actually equivalent to a penalty approach to the Lagrangian dual of the original SDP relaxation~\eqref{eq:map}.
For an sufficiently large parameter $\gamma$,
the simplified dual \eqref{eq:fastsdp_dual}
can be arbitrarily close to the original SDP relaxation \eqref{eq:map},
which however results in numerical problems.
In many cases, it is sufficient to restrict $\gamma$ to a large value
and obtain an approximate solution.
{
Note that how to choose the value of $\gamma$ in practice to make a good tradeoff between bound quality and convergence speed is
still an open question.
A potential strategy is to adjust $\gamma$ adaptively at each iteration of descent step.
}

The dual objective function $\mathrm{d}_\gamma(\cdot)$
is continuously differentiable but not necessarily twice differentiable,
and its gradient is given by
\begin{align}
\label{eq:dual_gradient}
\nabla \mathrm{d}_\gamma(\bu) = - \gamma \Phi \left[ \Pi_{\mathcal{S}^{n\!h\!+\!1}_+} \left(\bC(\bu)\right) \right] - \bb,
\end{align}
where $\Phi: \mathcal{S}^{nh+1} \rightarrow \mathbb{R}^m $
refers to the linear transformation $\Phi[\Omega] := [\langle \bB_1, \Omega\rangle, \cdots, \langle \bB_m, \Omega\rangle]^\T$.
Consequently,
we can solve the dual~\eqref{eq:fastsdp_dual} using quasi-N\-ew\-t\-o\-n algorithms
as they only require first-order derivatives during optimization.
At each descent step,
quasi-Newton methods update the dual variable $\bu$ as follows:
\begin{align}
\bu \leftarrow \bu + \rho \bH \nabla \mathrm{d}_{\gamma}(\bu),
\end{align}
where $\bH$ is the approximated inverse of Hessian matrix
updated by successive gradient vectors and
$1 \geq \rho \geq 0$ is the step size decided by line-search.

Another important property of the simplified dual problem
\eqref{eq:fastsdp_dual}
is that
{\em for any feasible dual variable $\bu$, the dual objective function
  value $\mathrm{d}_\gamma(\bu)$ is a lower-bound
to the minimum energy ${\min_{\bx \in \setZ^n}} \mathrm{E}(\bx)$}\ \cite{peng2013cpvr}.
This is beneficial for accelarating the \BB methods:
the bounding procedure can be stopped once the computed lower-bound is
not lower
than the current global upper-bound, and the corresponding subproblem can be pruned.

In the sequel, $\mathrm{d}^\star_\gamma(\setD)$ is defined as
the optimal value of the dual problem \eqref{eq:fastsdp_dual}
constructed \wrt $\gamma$ and $\setD \subseteq \setZ^n$.
In this paper, we further prove the following results for the tightness of the
lower-bound yielded by $\mathrm{d}^\star_\gamma(\setD)$ (see Appendix~\ref{app:2}):
\begin{proposition}
\label{prop1}
The following results hold:
($i$) $\forall \gamma \!>\! 0, \forall \setD_1 \!\subseteq\! \setD_2 \!\subseteq\! \setZ^n$,
we have $\mathrm{d}^\star_\gamma(\setD_1) \geq \mathrm{d}^\star_\gamma(\setD_2)$;
($ii$) $\forall \gamma \!>\! 0, \forall \mathcal{D} \!\subseteq\! \setZ^n$ with $\lvert {\mathcal{D}} \rvert = 1$,
we have $\mathrm{d}^\star_\gamma (\setD) = \min_{\bx \in \setD} \mathrm{E}(\bx)$.
\end{proposition}
The first result shows that the lower-bound is improved as the search space $\setD$ shrinks,
and the second means that the gap between the lower- and upper-bound
$\mathrm{E}(\bx)_{\bx \in \setD} - \mathrm{d}^\star_\gamma(\setD)$
converges to zero when the search space $\setD$ shrinks to a point (namely $\lvert \mathcal{D} \rvert = 1$).
In particular, if $|\setD| = 1$ and only contains
the MAP solution, $\arg \min_{\bx \in \setZ^n} \mathrm{E}(\bx)$,
$\mathrm{d}^\star_\gamma(\setD)$ equals to the global minimum energy.
These two properties are crucial when embedding the proposed bounding method into
branch-and-bound.

\begin{proposition}
\label{prop2}
(Optimality Guarantee of SDP Relaxation)
For any $\gamma > 0$, suppose $\{ \by_\gamma^\star, \bY_\gamma^\star \}$
and $\bu_\gamma^\star$ are the optimal/dual primal
variable obtained by \eqref{eq:fastsdp_primal_dual}.
If the rank of $(\Omega({\by_\gamma^\star, \bY_\gamma^\star}))$ is $1$, then
$\{ \by_\gamma^\star$, $\bY_\gamma^\star \}$ yields the exact MAP solution and
$\mathrm{d}_\gamma(\bu_\gamma^\star)$ is the minimum energy value.
\end{proposition}
The above
optimality guarantee of SDP relaxation
can be considered as a counterpart to Theorem~\ref{thm:lp} for the standard LP relaxation.
Both LP and SDP relaxations drop non-convex constraints and result in a simple convex relaxation:
the integer constraint is relaxed by LP relaxation and SDP relaxation drops the rank-$1$ constraint.
Note that Proposition~\ref{prop2} can be generalized to SDP relaxation %
with
linear constraints \eqref{eq:sdprelax_cons1}, \eqref{eq:sdprelax_cons2}.

Computationally  SDCut is significantly more scalable and efficient than general interior-point methods.
It is employed here to solve the SDP relaxation with linear constraints in \eqref{eq:map_sdcut_rank1_lcons}.
However, there are $\mathcal{O}(n^2h^2)$ number of
constraints \eqref{eq:sdprelax_cons3}, \eqref{eq:sdprelax_cons4} and the exponential number of constraints \eqref{eq:tricons}, \eqref{eq:cyccons}, \eqref{eq:owccons}.
Adding these constraints directly to SDCut is still impractical.
To this end, we proposed to find and add most violated constraints via cutting-plane,
which is discussed in the next section.

\subsection{Cutting-Plane: Searching and Adding Violated Constraints}
\label{eq:cutting-plane}

In practice, we find that most of the constraints \eqref{eq:sdprelax_cons3}, \eqref{eq:sdprelax_cons4}, \eqref{eq:tricons}, \eqref{eq:cyccons} and \eqref{eq:owccons}
are redundant after the initial bounding procedure shown in
Section~\ref{sec:initbounding},
which means that most of them are already satisfied after imposing
the \psd constraints and the linear constraints \eqref{eq:sdprelax_cons1}, \eqref{eq:sdprelax_cons2} and \eqref{eq:sdprelax_cons5}.
Therefore, cutting-plane methods can be adopted to search violated constraints in
\eqref{eq:sdprelax_cons3}, \eqref{eq:sdprelax_cons4}, \eqref{eq:tricons}, \eqref{eq:cyccons} and \eqref{eq:owccons},
and add them to SDP relaxation at each iteration.
Adding these primal constraints is equivalent to adding new variables to the dual problem~\eqref{eq:fastsdp_dual}.

{
Violated non-negativity~\eqref{eq:sdprelax_cons3}, edge
marginalization~\eqref{eq:sdprelax_cons4} constraints and triangular inequalities~\eqref{eq:tricons}
are found by enumeration.
Most violated cycle inequalities~\eqref{eq:cyccons} are found using the separation procedure~\cite{deza1996geometry}.
When there is no violated cycle inequality (up to certain precision),
violated odd-wheel inequalities~\eqref{eq:owccons} are searched and added using the separation procedure shown in \cite{deza1996geometry}.
}

The procedure of the proposed SDP bounding procedure, including
initial bounding (Section~\ref{sec:initbounding}) and cutting-plane
(Section~\ref{eq:cutting-plane}) are summarized in Algorithm~\ref{alg:bound}.
At each iteration of cutting-plane, the dual problem \eqref{eq:fastsdp_dual} is c\-on\-s\-tr\-u\-c\-t\-ed based
on three inputs: the search space $\setD$ (for reparameterization),
linear constraints set $\Psi$ and the penalty parameter $\gamma$.

\subsection{Computational Complexity}
\label{sec:complexity}

L-BFGS-B~\cite{Zhu94lbfgsb} is used as the implementation of quasi-N\-ew\-t\-o\-n algorithms for SDCut.
At each iteration,
L-BFGS-B itself requires only $\mathcal{O}(m)$ arithmetic operations and memory requirement,
so SDCut scales linearly in the number of constraints $m$.
The main computational burden of SDCut is the eigen-decomposition at each gradient-descent step
to compute the dual objective function~\eqref{eq:dual_obj} and its gradient~\eqref{eq:dual_gradient}.
The eigen-solver used
in this paper is the {\tt DSYEVR}
routine in {\tt LAPACK} (an implementation of Relatively Robust Representations),
which has the computational complexity of $\mathcal{O}(n^3h^3)$ and the memory requirement of $\mathcal{O}(n^2h^2)$.

Finding the most violated non-negativity constraint~\eqref{eq:sdprelax_cons3}
and edge marginalization constraint~\eqref{eq:sdprelax_cons4} requires $\mathcal{O}(n^2h^2)$ arithmetic operations,
while finding violated cycle inequality needs $\mathcal{O}(n^3h^3)$ arithmetic operations.
To save time, only a few center node $\pi_r$ (see Equation~\eqref{eq:owccons}) is considered for odd-wheel inequalities in each cutting-plane iteration.

Therefore the overall computational complexity of the proposed SDP method is $\mathcal{O}(n^3h^3)$ at each descent iteration
($m$ is less than $\mathcal{O}(n^3h^3)$, so it is omitted),
and the corresponding memory requirement is $\mathcal{O}(n^2h^2)$.
This result de\-monstrates that our approach is much more efficient than the classic interior-point methods,
which needs $\mathcal{O}(n^6h^6)$ arithmetic operations.

\subsection{Strategies to Speed Up}
\label{sec:sdp_speedup}

Although the proposed SDP approximation method already provides a
tight lower-bound efficiently,
we proposed a few techniques to further speed up the bounding
procedure of \BB.

{
\noindent
{\bf Pre-Processing (Model Reduction)}
Several methods have been proposed to identify persistency for binary models \cite{rother2007optimizing},
{ Potts models~\cite{kovtun2011sufficient,swoboda2013partial} } and
general multi-label models
\cite{kohli2008partial,kovtun2011sufficient,windheuser2012generalized,swoboda2014partial,shekhovtsov2014maximum,shekhovtsov2015maximum}.
Persistencies refer to states of variables which are proved to be non-optimal or belong to at least one optimal solution.
Previous work has employed persistency to reduce the model size
before applying approximate~\cite{alahari2008reduce} and exact~\cite{kappes2013towards} solvers.
For binary models,
all variables assigned as integers by QPBO~\cite{rother2007optimizing} are persistent.

{\em Without affecting the property of global convergence of \BB},
QPBO is performed in each bounding procedure,
and then the SDP bounding procedure is applied on the reduced model.
In general, the
computation time of QPBO is negligible compared to that of the SDP procedure. Because the
size of the resulted
SDP is smaller, the total computational time is reduced as well.
Note that this partial optimality also holds for
TRW message passing and other methods equivalent to the standard LP relaxation for binary graphs.
We choose QPBO here for its fast computation.

\noindent
{\bf Stop Conditions}
As discussed in Section~\ref{sec:initbounding},
the objective value of \eqref{eq:fastsdp_dual}, calculated at each descent step,
can be cast as a lower-bound to the minimum energy.
For faster computation,
the bounding process is
stopped before convergence if any of the following { conditions} is met (see Algorithm~\ref{alg:bound}):%
\begin{enumerate}
\item The lower-bound $\mathrm{lb}$ is already greater than the global upper-bound $\mathrm{gub}$,
  and thus this subproblem is pruned.
\item The improvement of the lower-bound $\mathrm{lb}$ between consecutive steps is smaller than the specified tolerance.
\end{enumerate}

\noindent
{\bf Warm Start}
Another strategy used in the bounding procedure to speed up the computation is ``warm start",
which is performed in two levels: between different iterations of cutting-planes and bounding procedures.
\begin{enumerate}
\item In an iteration of cutting-plane, the dual variables are initialized by the results of the previous iteration.
For the new constraints, the corresponding dual variables are filled with zeros.
\item The final working set of active constraints and the corresponding dual variables of the first bounding iteration
are used to initialize those of subsequent bounding iterations ($\psi_{\mathrm{init}}$ and $\bu_{\mathrm{init}}$ in Algorithm~\ref{alg:bound}),
which speed up the convergence speed of L-BFGS-B.
The initial working set $\psi_{\mathrm{init}}$ of the first bounding iteration is set to contain constraints \eqref{eq:sdprelax_cons1},\eqref{eq:sdprelax_cons2} and \eqref{eq:sdprelax_cons5},
and $\bu_{\mathrm{init}}$ is set to zeros.
\end{enumerate}

\noindent
{\bf Removing Inactive Constraints}
One disadvantage of cutting plane is that the size of the subproblems
(the number of primal linear constraints, and correspondingly the number of dual variables)
keeps increasing from iteration to iteration,

Since traditionally new constraints are added in each r\-ou\-n\-d but are never removed.
The strategy of pruning inactive constraints has been employed
to address this issue~\cite{topkis1982cutting,joachims2009cutting,horst2013global}
in cutting-plane methods.
In this work, we use a simple rule to decide which constraint should be dropped from the working set:
those corresponding to zero dual variables and not violated by the current primal solution.
Note that those constraints discarded by the current iteration
may be re-activated in subsequent cutting-plane iterations.

\noindent
{\bf Post-Processing (Stochastic Local Search)}
\label{sec:ricm}

When the dual optimization is terminated,
an approximate solution $\bx_{\text{appx}} \in \setZ^n$ is calculated by rounding
and an upper-bound $\mathrm{E}(\bx_{\text{appx}})$ to the minimum
energy $\min_{\bx \in \setZ^n} \mathrm{E}(\bx)$ is obtained.
Local (greedy) search methods can be applied to further tighten the upper-bound.
What we use here is `repeated iterated conditional modes' (refer to as RICM):
\begin{enumerate}
  \item
 Run Iterated Conditional Modes (ICM)\ \cite{besag1986statistical} with $\bx_{\text{appx}}$
as the initial point to find a local optimal solution
$\bx_{\text{local}}$.
      Compute the upper-bound $\mathrm{E}(\bx_{\text{local}})$.
    \item
      Randomly flip a portion (around $5\%$) of assignments in the initial solution $\bx_{\text{appx}}$.
    \item
      Repeat the steps 1 and 2 several times and
    select the local optimal solution $\bx_{\text{local}}$ with the lowest upper-bound as the new approximate solution.
\end{enumerate}

    Note that ICM can be considered as a greedy coordinate descent method.
The solution quality is guaranteed to be improved or at least not
worse than the initial point. Furthermore, as ICM depends highly on
the initialization, we repeat ICM several times to overcome this
problem.

}

\section{Experiments}
\label{sec:exp}

{

The proposed method is evaluated in two
forms: one with cutting-planes (refer to as {SDBC}) and
one without
(refer to as {SDBB}).
Both of the SDP bounding methods in SDBB and SDBC
solve the SDP relaxation~\eqref{eq:map} approximately based on the simplified dual formulation~\eqref{eq:fastsdp_dual}.
The only difference is that SDBC imposes all linear constraints listed in \eqref{eq:map_cons}\footnote{In the following experiments,
  we find that odd-wheel inequalities are only effective on the modularity clustering models,
therefore this class of constraints is not considered for other models},
while SDBB only considers constraints \eqref{eq:sdprelax_cons1}, \eqref{eq:sdprelax_cons2} and \eqref{eq:sdprelax_cons5}.
The following algorithms for MAP-MRF inference are compared against ours.

 1) As a baseline, the lower-bound produced by TRWS~\cite{kolmogorov2006convergent} is  compared with SDBC.

 2)
Sontag \etal~\cite{sontag2008,sontag2012efficiently} proposed to tighten the LP relaxation to MAP problems by
incrementally introducing cluster consistency constraints, \eg, triplet clusters~\cite{sontag2008} (refer to as {MPLP-CP-v1})
or long cycles~\cite{sontag2012efficiently} (refer to as {MPLP-CP-v2}).
The code is obtained from the author's website. %
At each round, $50$ additional triplets are added to MPLP-CP-v1 and $50\!\sim\!200$ triplets plus $50\!\sim\!200$ cycles
are added to MPLP-CP-v2.
MPLP-CP-v1 and MPLP-CP-v2 provide lower-bounds to the global optimal energy value, which can be used to validate the exactness of a solution.
For SDP relaxation methods,
we compared our method with SDPT3, which represents one of the state-of-the-art
implementations of interior-point methods.

3)
Our method are also compared with methods based on graph cuts~\cite{kolmogorov2007minimizing,gorelick2014a} for binary non-submodular pairwise energies.
QPBO~\cite{kolmogorov2007minimizing} can be used to exactly solve problems with a small fraction of non-submodular terms.
We also use QPBO to reduce the model before applying other methods.
Local submodular approximation (LSA)~\cite{gorelick2014a}
iteratively approximates non-submodular energies non-linearly based on trust region (refer to as LSA-TR) or auxiliary function principles.

4)
Several \BB methods for energy minimization are also evaluated.
Sun \etal~\cite{Sun2012MRFBB2} proposed a \BB method (refer to as {MPLP-BB})
using MPLP to upper-bound the MAP value at each branch.
The dual variables of MPLP at each branch are fixed,
and a data structure, called range maximum query, is pre\--con\-s\-tr\-uct\-ed to
speed up the bound calculation significantly.
The parameters of MPLP-BB are set to the default values,
except that the number of initial MPLP iterations is set to $1000$.

DAOOPT~\cite{otten2012anytime} implements the And/Or Branch and B\-ou\-n\-d paradigm,
which also uses mini-bucket heuristic for pruning search space and limited discrepancy search (LDS) to find an initial solution quickly.
It achieves the best performance in the PIC2011 Challenge~\cite{pic2011}.
In this paper, the memory limit of DAOOPT is set to $8$GB and the discrepancy limit of LDS is set to $2$.
The commercial integer linear program solver, IBM CPLEX~\cite{cplex} (denoted by ILP), is also evaluated in the experiments.
Some specialized algorithms~\cite{kernighan1970a,bonato2014lifting,kappes2011globally,kappes2013higher} are also investigated for some particular data sets.

To perform a fair comparison,
all the implementations of compared methods are obtained from the author's website or from the OpenGM2 toolbox~\cite{opengm2} and tested
on one core of a $2.7$GHz Intel CPU with $4$GB free memory, except for DAOOPT ($10$GB) and ILP ($40$GB).
Note that some experimental results of competitive methods are obtained from \cite{opengm2}, which is evaluated on a faster Intel CPU ($3.2$GHz).
All the methods are performed with the same procedures of pre-processing (model reduction by QPBO) and post-processing (stochastic local research by ICM).
The primary results (Table~\ref{tab:deconvolution}, \ref{tab:objdet}, \ref{tab:chinchar} and \ref{tab:moducluster}) are shown in a format similar to \cite{kappes2013comparative},
which demonstrates the upper-bound (energy value of the final integer solution) and lower-bound (to the optimal energy value),
averaged over all instances of each problem.
For an algorithm, the number of instances for which it achieves the best upper-/lower-bound among all algorithms
(denoted as $\#$Best-ub and $\#$Best-lb respectively)
and the number of instances to which it gives the exact solution (denoted as $\#$Exact)
are also demonstrated for better comparison.
As in \cite{kappes2013comparative}, a solution is considered to be exact if the difference between the lower-bound and upper-bound is less than $10^{-5}$
in terms of absolute value or less than $10^{-8}$ in terms of relative value.

The implementation of our algorithm is written in matlab and the implementation
of other methods are mainly implemented in C/C++.
We would expect a further improvement in the speed of our method if it were
implemented in optimized C/C++.

\subsection{Performance Analysis}

\begin{figure}
\centering
\includegraphics[type=pdf,ext=.pdf,read=.pdf,width=0.4\textwidth]{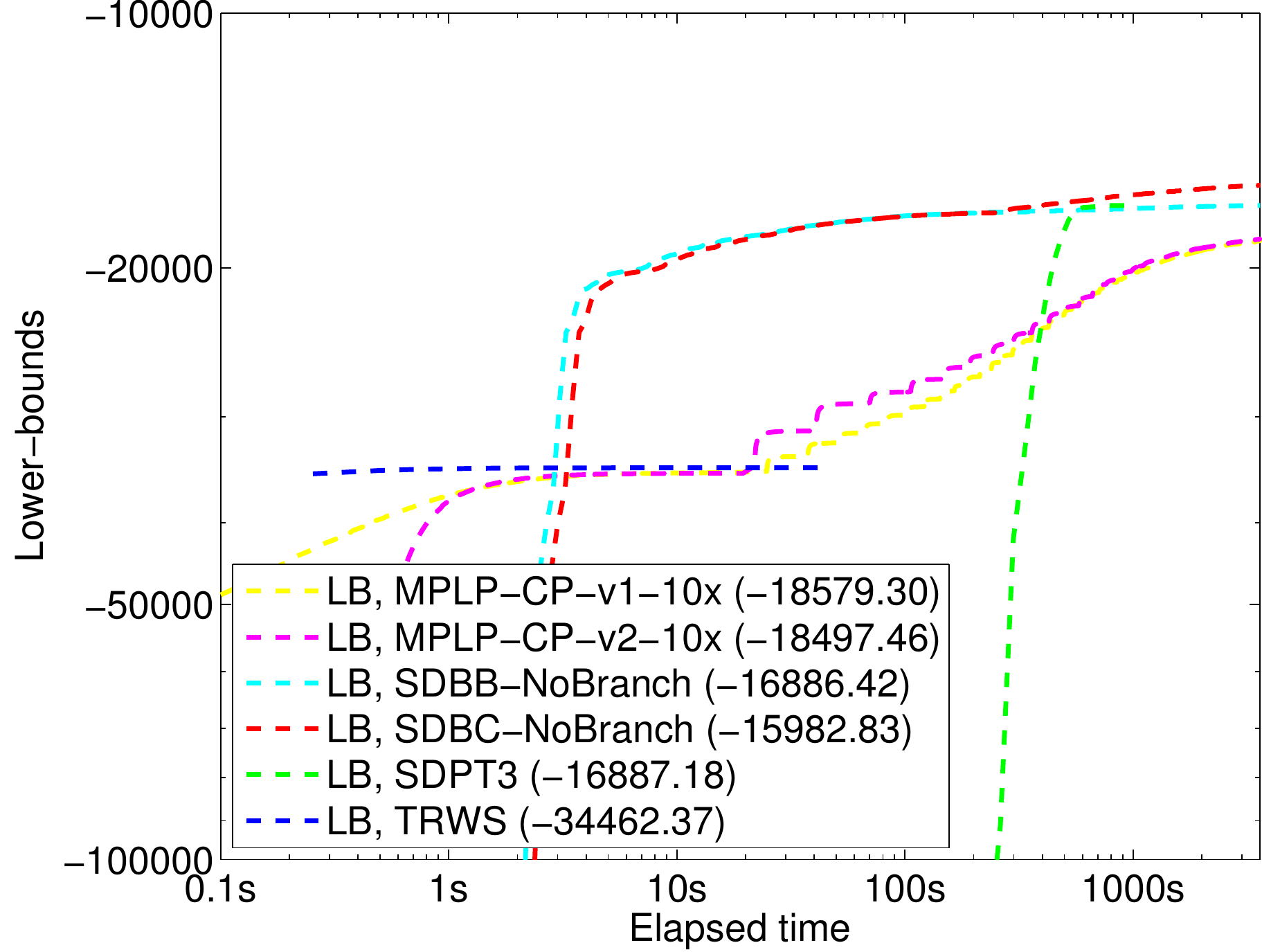}%
\caption{
{
Lower-bounds given by different algorithms on the model ``deer$\_$rescaled$\_$0034.K15.F100" ($60$ variables, $15$ states per variable) in PIC2011.
          SDBC is performed without branching (SDBC-NoBranch) or without both branching and cutting-plane (SDBB-NoBranch).
          The lower-bounds provided by LP-based methods (TRWS and MPLP-CP) are worse than SDP-based methods (SDPT3 and SDBC).
          SDBB-NoBranch produces a lower-bound very close to the interior-point method SDPT3.
          For SDBC itself, the additional constraints added through cutting-plane gives a $5\%$ increase in lower-bound.}
}
\label{fig:compare_lb}
\end{figure}

\begin{figure*}[t!]
\vspace{-0.0cm}
\centering
\subfloat[{fully connected, $\omega = 0.4$}]{
\centering \includegraphics[type=pdf,ext=.pdf,read=.pdf,width=0.32\textwidth]{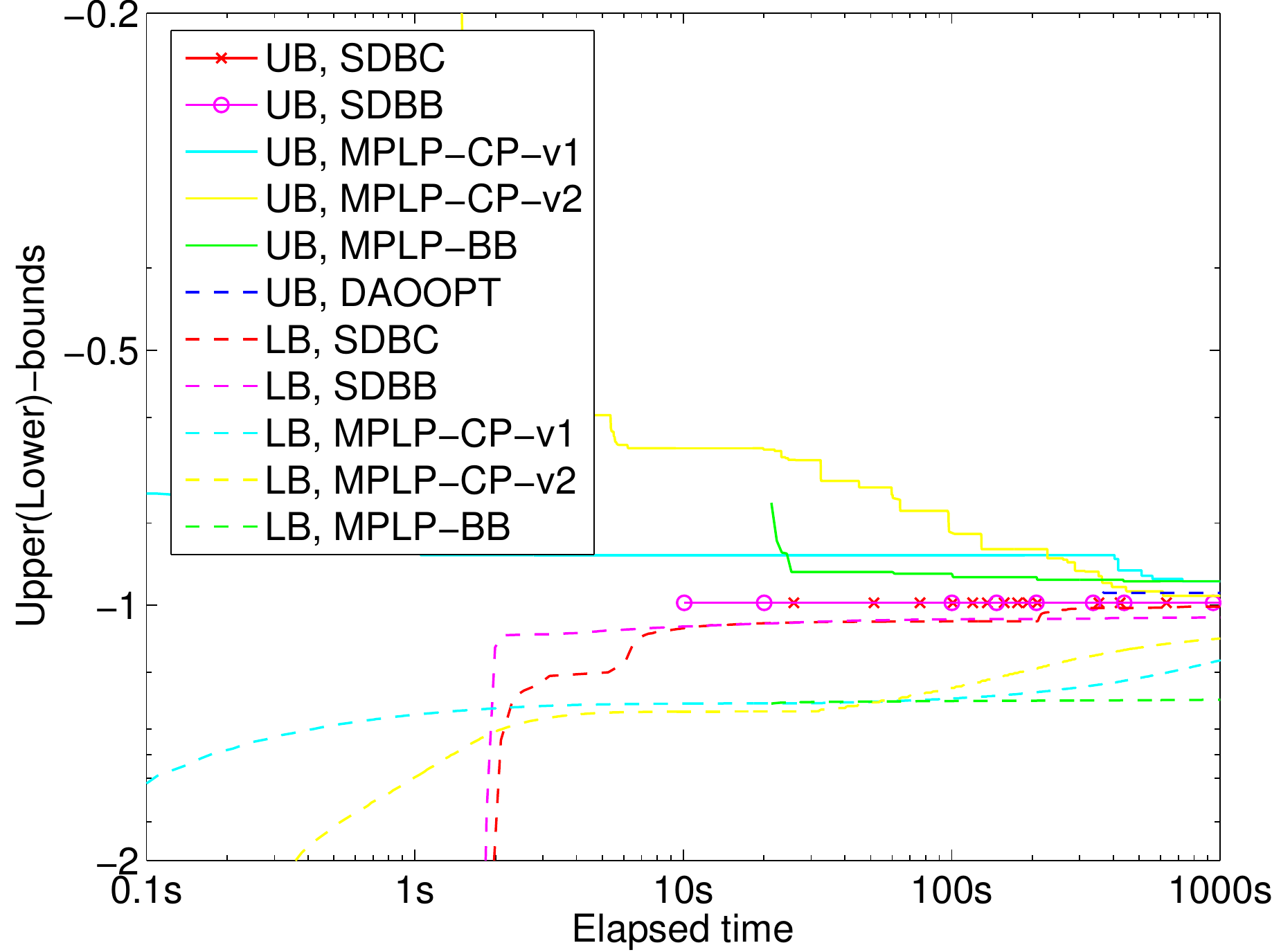}}
\subfloat[{fully connected, $\omega = 0.2$}]{
\centering \includegraphics[type=pdf,ext=.pdf,read=.pdf,width=0.32\textwidth]{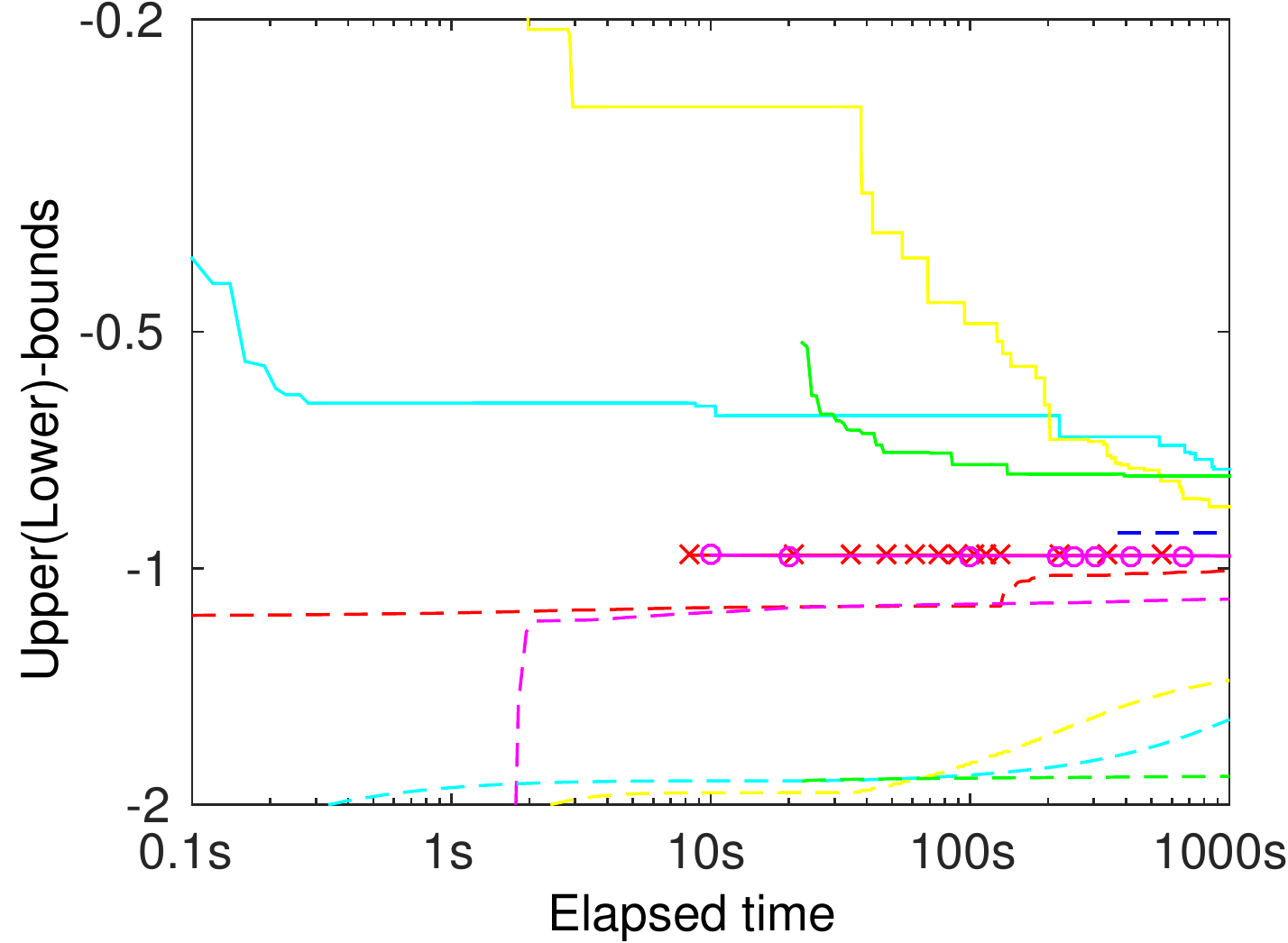}}
\subfloat[{fully connected, $\omega = 0.1$}]{
\centering \includegraphics[type=pdf,ext=.pdf,read=.pdf,width=0.32\textwidth]{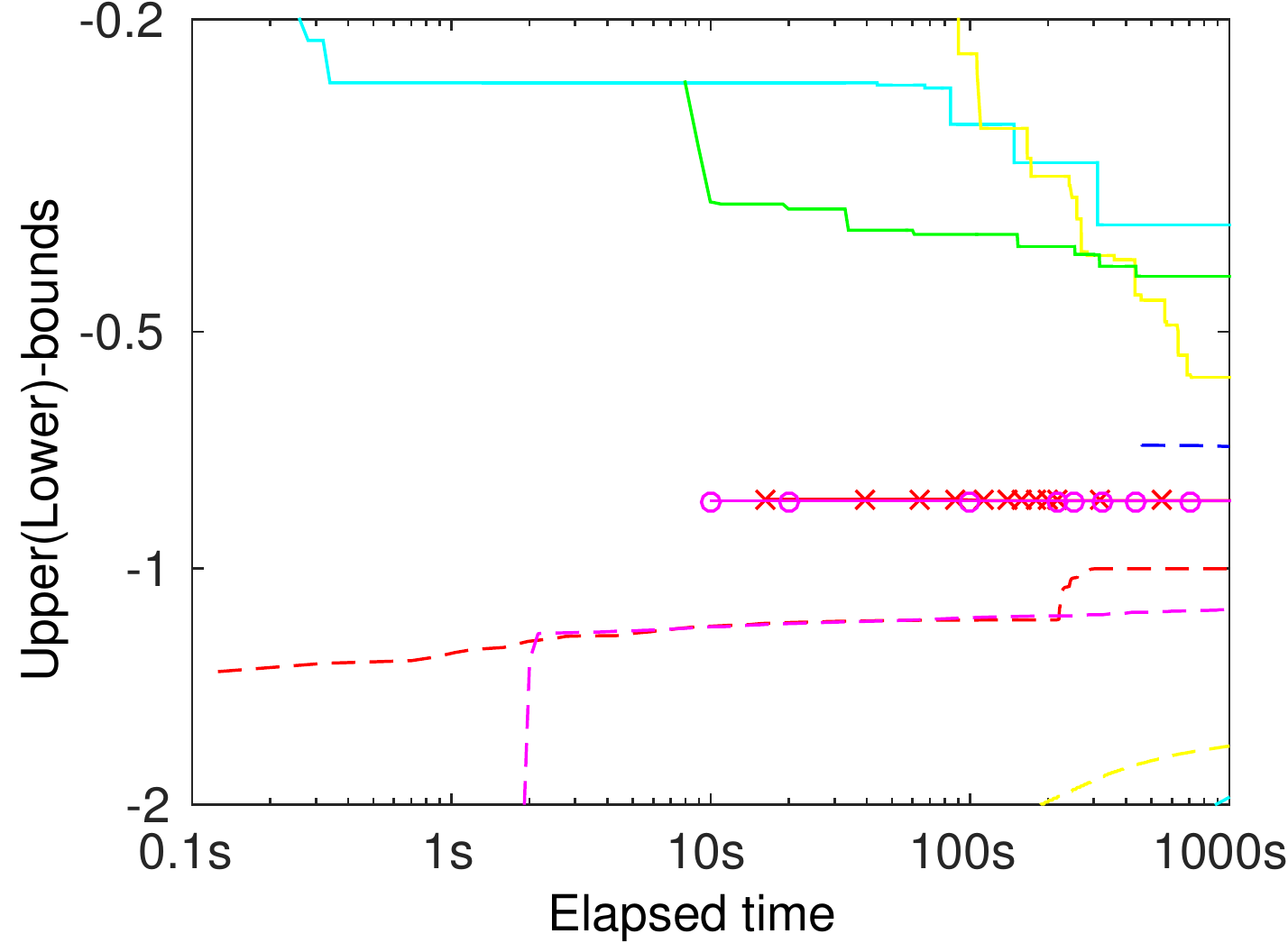}}\\
\subfloat[{$\kappa = 6.56$, $\omega = 0.25$}]{
\centering \includegraphics[type=pdf,ext=.pdf,read=.pdf,width=0.32\textwidth]{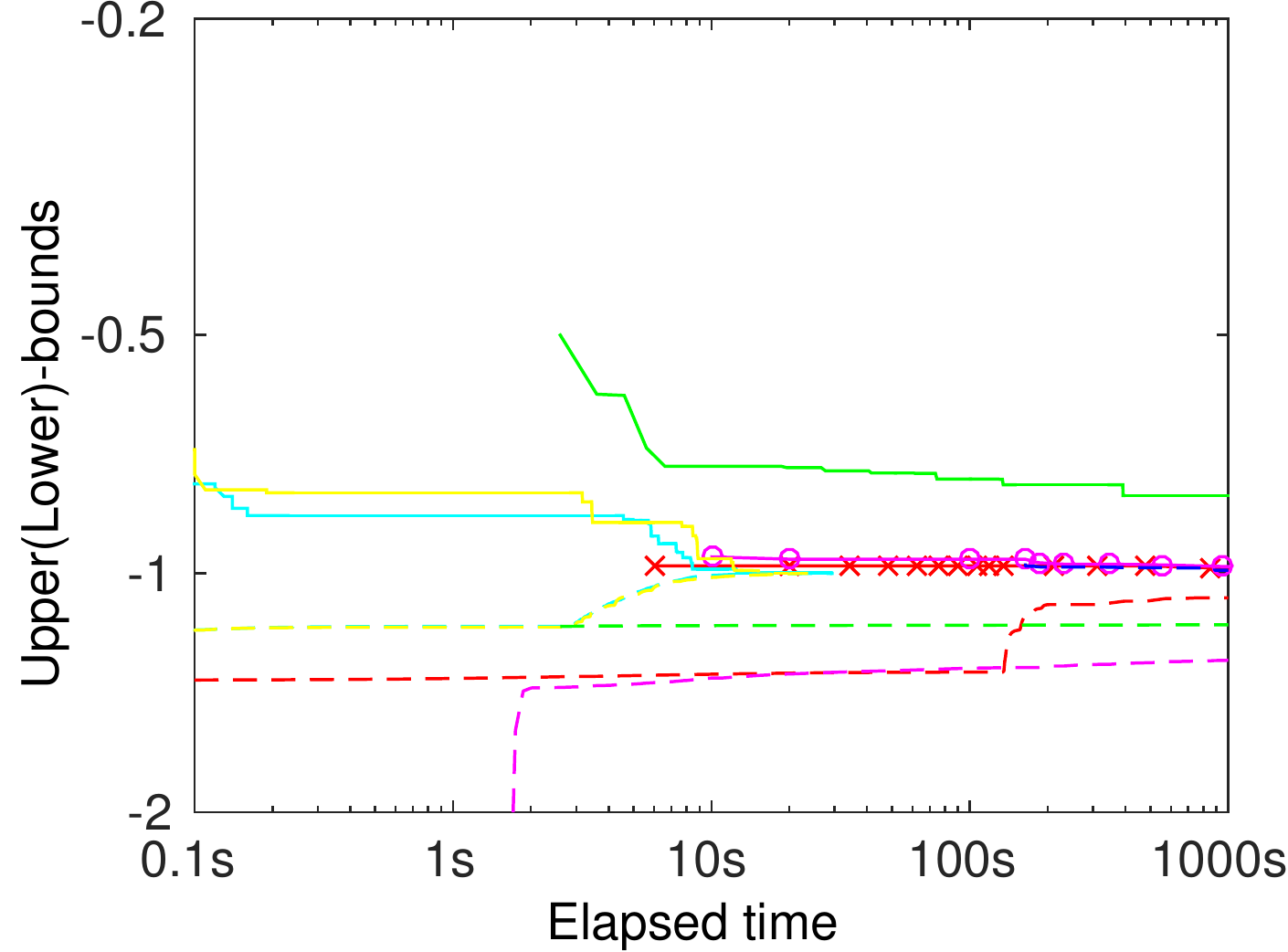}}
\subfloat[{$\kappa = 17.06$, $\omega = 0.25$}]{
\centering \includegraphics[type=pdf,ext=.pdf,read=.pdf,width=0.32\textwidth]{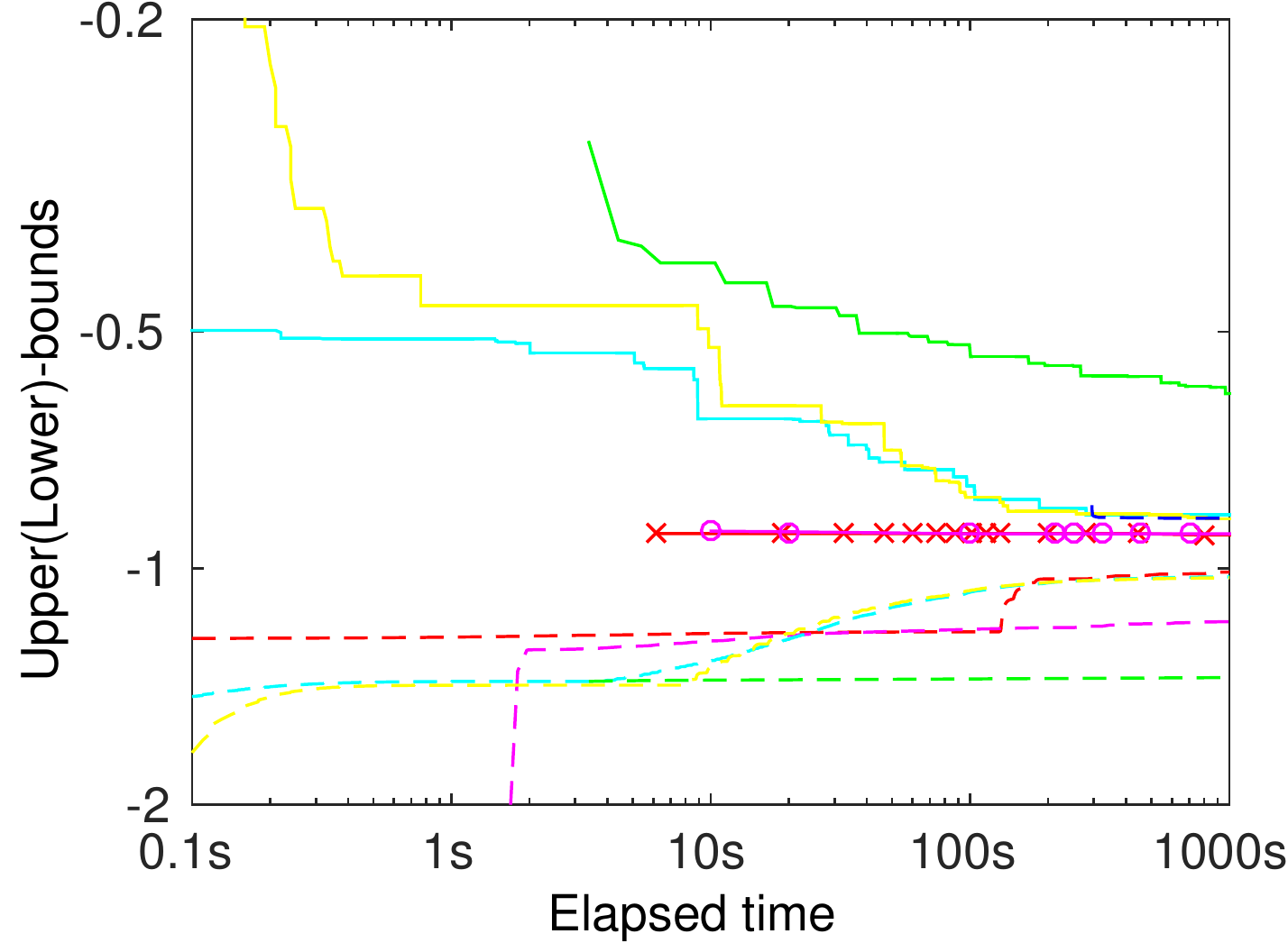}}
\subfloat[{$\kappa = 59.06$, $\omega = 0.25$}]{
\centering \includegraphics[type=pdf,ext=.pdf,read=.pdf,width=0.32\textwidth]{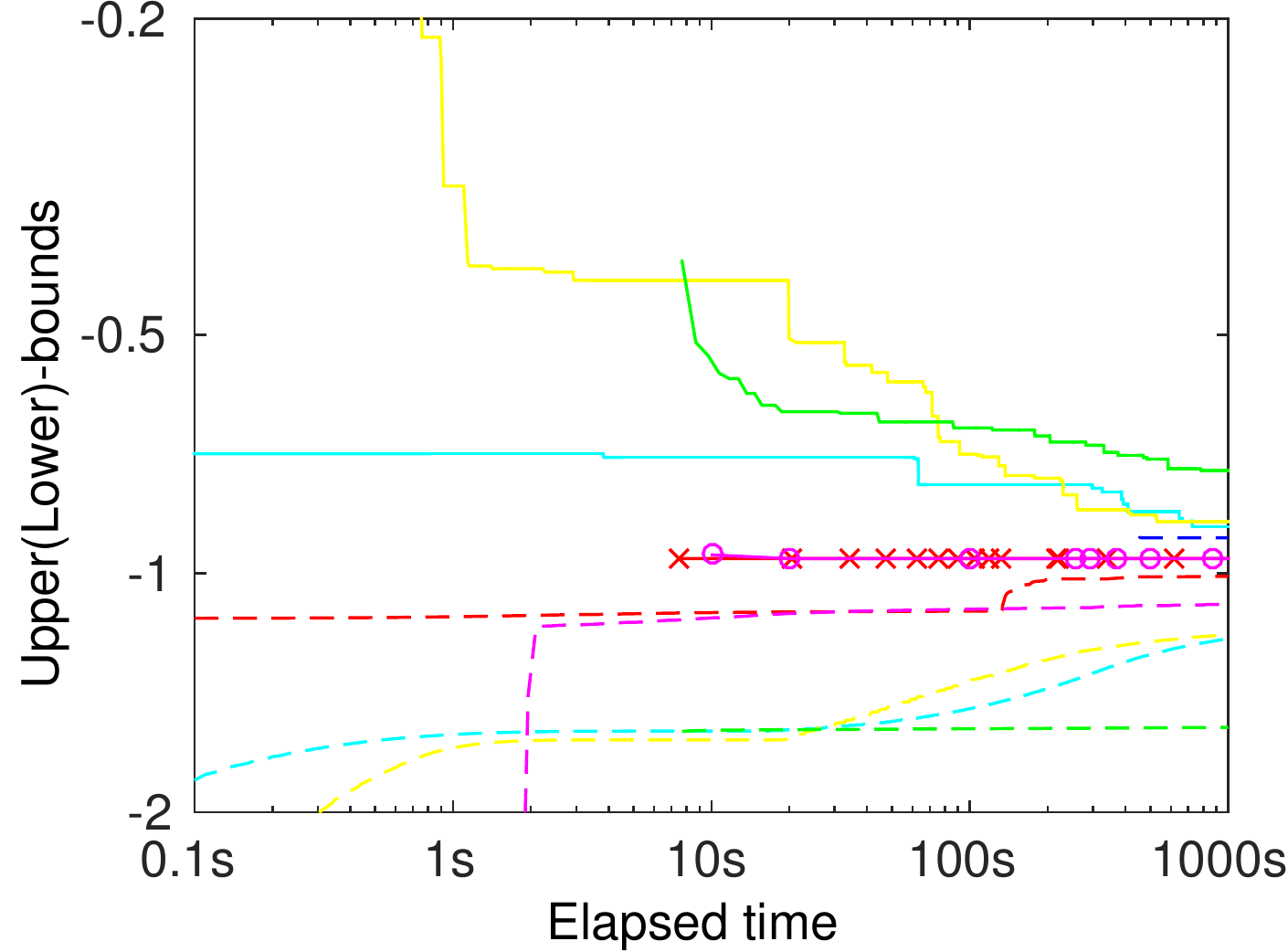}}
   \caption{Upper-/lower-bound versus running time on the synthetic MAP-MRF inference problems.
            All the undirected graph models in this experiment have $64$ nodes and $10$ states per node, and the energy functions are randomly generated.
            The upper/lower bound is normalized by dividing them by the best lower-bound computed within $20$ minutes.
            In (a)-(c), the unary strength decreases from $0.4$ to $0.1$.
            When the unary strength parameter $\omega$ is large (\eg, $\omega \!=\! 0.4$),
            all methods obtain near optimal upper-bounds ($<\!-0.90$) and the superiority of SDBC is not significant.
						SDBC does perform
						significantly better than others for smaller unary strengths (\eg, $\omega \!=\! 0.1$).
            In (d)-(f), the connectivity $\kappa$ changes from $6.56$ to $59.06$. For sparse graphs (\eg, $\kappa\!=\!6.56$), MPLP-CP-v1 and MPLP-CP-v2 converge quickly to the optimal and performs better than SDBC.
            However for dense graph models (\eg, $\kappa\!=\!59.06$), SDBC yields better
						upper and lower bounds.
            }
\label{fig:synthetic}
\end{figure*}

{\bf Relaxation Tightness}
The SDP bounding procedure embedded in SDBC
is compared with state-of-the-art SDP/LP relaxation methods.
SDPT3~\cite{tutuncu2003solving}, an implementation of interior point methods,
is used to accurately solve the SDP relaxation with
linear constraints \eqref{eq:sdprelax_cons1}, \eqref{eq:sdprelax_cons2} and \eqref{eq:sdprelax_cons5}.
Note that the other classes of tightening constraints are difficult to be imposed with SDPT3,
due to the poor scalability of interior-point methods in
the number of linear constraints.
MPLP-CP-v1 and MPLP-CP-v2 tighten the standard LP relaxation (over local marginal polytope $\mathcal{M}_L(\mathcal{G},\setZ)$)
by incrementally adding high-order consistency constraints.

In Fig.~\ref{fig:compare_lb}, the performance of these relaxation methods are compared
on the lower-bounds achieved on a model
in the PIC2011, ``deer$\_$rescaled$\_$0034.K15.F100", which is a fully-connected graph with $60$ nodes and $15$ states per node.
In this experiment, SDBC is performed without branching, which means there is only one bounding procedure.
Furthermore, to show the effectiveness of cutting-plane, SDBC is performed in two forms: with (SDBC-NoBranch) or without cutting-plane (SDBB-NoBranch).

From the results in Fig.~\ref{fig:compare_lb}, we can see that:
\begin{enumerate}
  \item
 SDBB-NoBranch achieves a final lower-bound very similar to that produced by SDPT3 (the difference is only $0.1\%$),
   but is significantly faster than SDPT3. %
   Note that SDBB-NoBranch and SDPT3 solve the same SDP problem, but
   approximately and exactly  respectively.
   As more linear constraints are imposed, SDBC-NoBranch achieves better lower-bound than SDBB-NoBranch and SDPT3 (over $3\%$).

 \item
   Both SDBB-NoBranch and SDBC-NoBranch yield a tig\-h\-ter lower-bound
   than TRWS, MPLP-CP-v1 and MPLP-CP-v2,
   which shows that the \psd constraint is indeed effective.
 \item
Lower bounds of SDBC-NoBranch and SDBB-NoBranch converge after certain iterations.
   To further improve lower bounds, we need to embed the SDP bounding procedure into \BC.
\end{enumerate}

\noindent
{\bf Connectivity and Unary Strength}
We now compare the performance of SDBC, SDBB and other MAP-MRF inference methods when applied to a range of
synthetic problems generated with varying parameters.
We consider two factors which affect the difficulty of MAP-MRF inference problems: connectivity (\ie, the average number of neighbours per node)
and unary
potential
strength (compared to pairwise potentials).
Higher connectivity and smaller unary
potential
strength increase the difficulty of MAP-MRF inference problems,
and degrade the performance of some existing algorithms (\eg, QPBO and belief propagation~\cite{kim2011hybrid}).

In this experiment, all the synthetic MRF models have $64$ variables and $10$ states per variable.
The pairwise energies are sampled from the Gaussian distribution $\theta_{pq}(i,j) \!\thicksim\! \mathcal{N}(0,1)$,
and the unary energies are sampled from $\theta_{p}(i) \!\thicksim\! \mathcal{N}(0,\omega^2)$,
where the parameter $\omega$ controls the unary strength: small $\omega$ corresponds to weak unary strength, and vice versa.
$\kappa$ refers to the connectivity of a graph, which is the average
number of linked neighbours per node.

The results are shown in Fig.~\ref{fig:synthetic}, in which
the curve of lower-bounds is the concatenation of two parts:
the lower-bounds at the gradient-descent steps within the first bounding iteration and the lower-bounds for the subsequent bounding iterations.
In the first bounding iteration of SDBB, multiple rounding procedures are conducted at $10$, $20$ and $100$ seconds respectively.

In Fig.~\ref{fig:synthetic} (a)-(c), fully-connected graphs ($\kappa = 63$) are generated and the parameter $\omega$ is set to $\{0.4, 0.2, 0.1\}$ respectively.
SDBC and SDBB achieve the best upper/lower bounds in most  cases.
Meanwhile, SDBC yields better lower-bound than that of SDBB.

The performance of all methods becomes worse when the parameter $\omega$ decreases.
However, the
detrimental impact on SDBC is the smallest amongst
all the evaluated algorithms.
The upper-bound of SDBC drops from $0.993$ ($\omega = 0.4$) to $0.820$ ($\omega = 0.1$),
while the upper-bound of the second best
method
decreases from $0.975$ to $0.700$.
The difference on the lower-bounds is more significant:
at $\omega = 0.4$, the gap between SDBC and the second best on the lower-bound is only $0.087$;
while the gap increases to $0.89$ for $\omega = 0.1$.

In Fig.~\ref{fig:synthetic} (d)-(e), $\omega$ is fixed to $0.25$ and the connectivity $\kappa$ is set to $\{6.56, 17.06, 59.06 \}$ respectively.
SDBC and SDBB perform slightly worse than MPLP-CP-v1 and MPLP-CP-v2
in terms of both upper and lower bounds
when $\kappa = 6.56$.
For denser graphs ($\kappa \in \{ 17.06, 59.06 \}$), however, SDBC produces the best lower-/upper-bounds amongst all competing methods.
{
We also compare SDBC and MPLP-CP-v1 on a larger sparse model from \cite{liu2015crf}, which contains $768$ variables, $8$ states per variable and $2252$ edges.
The upper-/lower-bounds of MPLP-CP-v1 converge quickly to the global optimal solution ($-565.23$) using $0.8$ second, as it can benefit from the sparse graph structure.
While SDBC scales poorer than MPLP-CP-v1 on this type of sparse models:
it obtains a non-global-optimal solution ($-567.92$) using around $2$-hour runtime.
Note that the focus of the proposed SDP approaches is on the dense problems with non-submodular terms
and/or weak unary terms (See \cite{wang2015efficient} for the efforts to improve the scalability of SDP algorithms for MAP inference).}

Based on the experiments on this synthetic data,
we may conclude that {\em SDBC and SDBB perform better than state-of-the-art
for MRFs with high connectivity or weak unary potentials}.

\subsection{Image Denoising and Deconvolution}

The binary submodular MRF model for image denoising is generated using the UGM\footnote{\url{http://www.di.ens.fr/~mschmidt/Software/UGM.html}} code.
As the energy provided by the graph-cut algorithm for such MRF models is the exact MAP value,
it is used to compare with ours to validate whether the proposed \BC method
converges
to the exact MAP solution.
As shown in Fig.~\ref{fig:denoise}, the result of SDBC is indeed the same as that of the graph-cut algorithm.

\begin{figure}
\subfloat[{\scriptsize Ground Truth}]{
\centering \includegraphics[type=pdf,ext=.pdf,read=.pdf,width=0.11\textwidth]{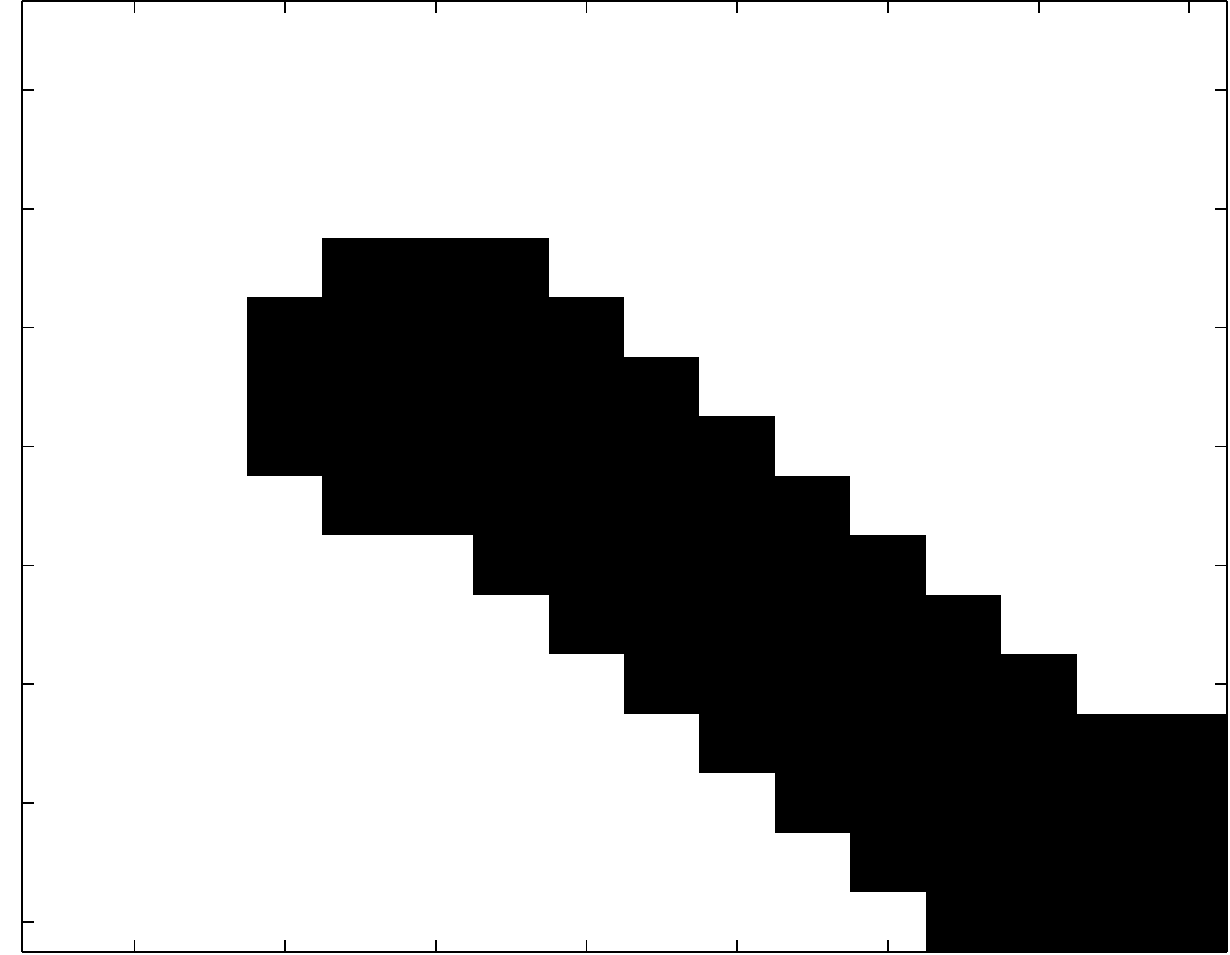}}
\subfloat[{\scriptsize Input}]{
\centering \includegraphics[type=pdf,ext=.pdf,read=.pdf,width=0.11\textwidth]{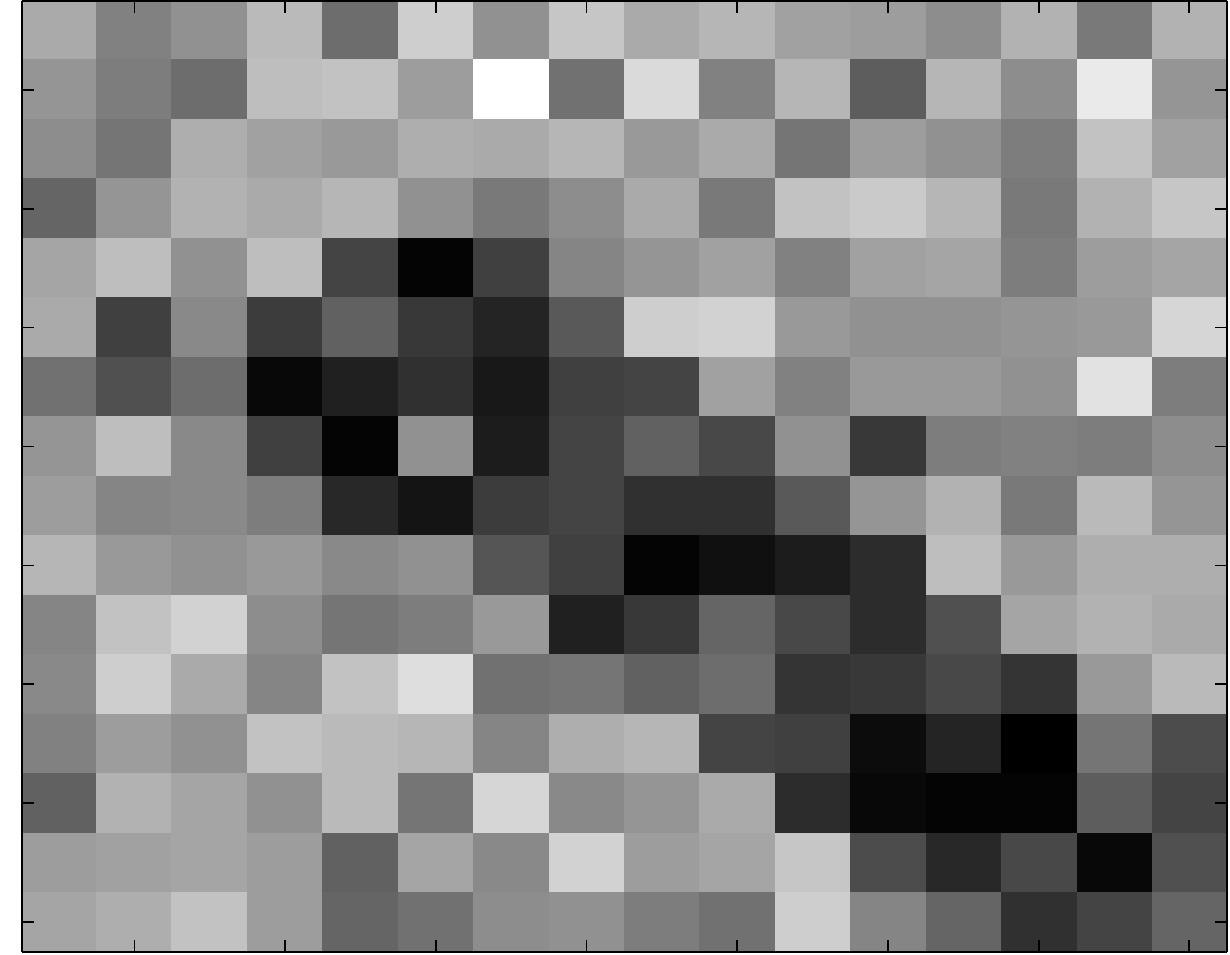}}
\subfloat[{\scriptsize Graph-cut}]{
\centering \includegraphics[type=pdf,ext=.pdf,read=.pdf,width=0.11\textwidth]{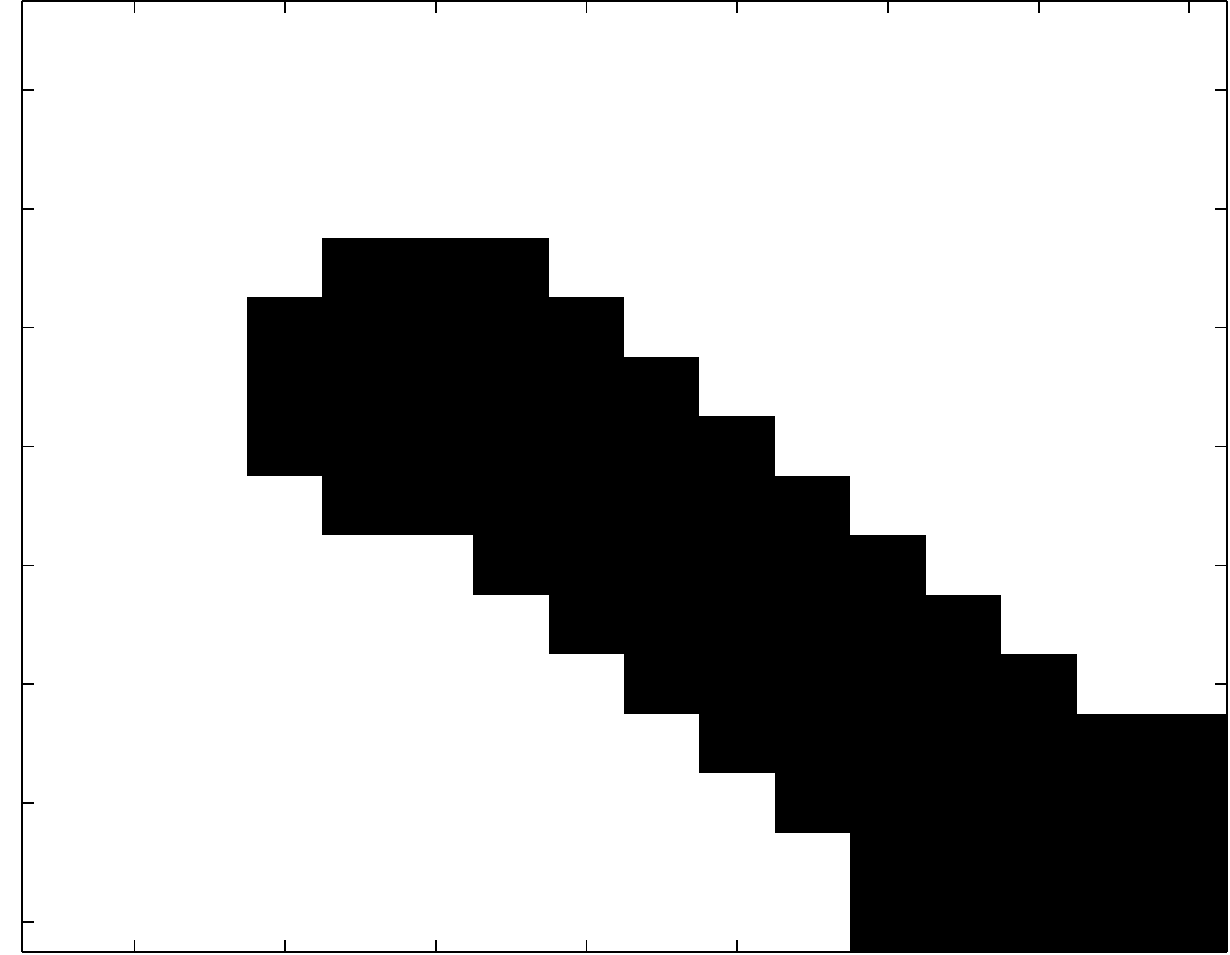}}
\subfloat[{\scriptsize SDBC}]{
\centering \includegraphics[type=pdf,ext=.pdf,read=.pdf,width=0.11\textwidth]{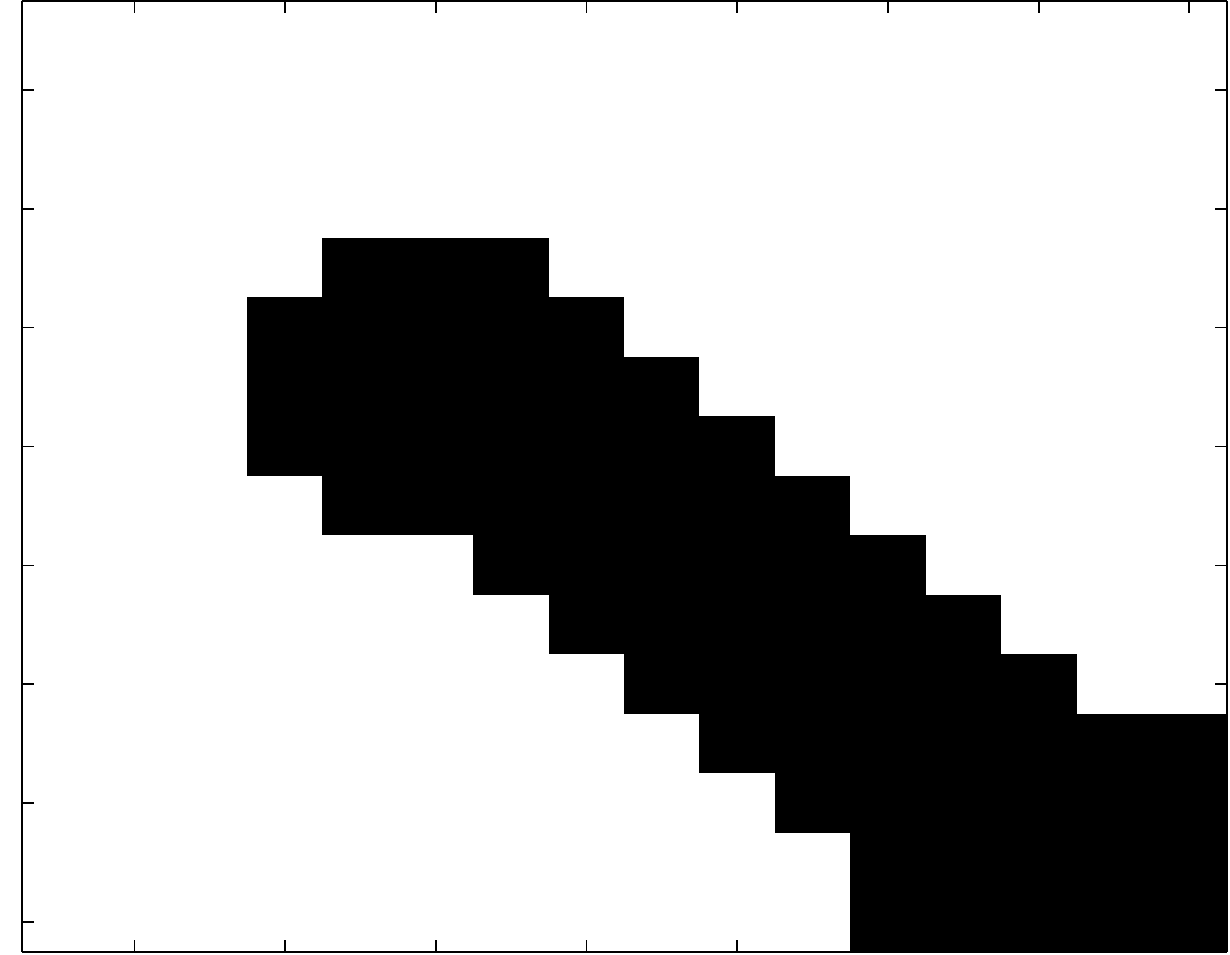}}
\caption{\footnotesize
{
Image denoising. This experiment is used to validate whether SDBC converges to the globally optimal solution.
         The $16\!\times\!16$ input image (b) is generated by adding random Gaussian noise to all pixels of the ground truth image (a).
         The con\-s\-tr\-u\-c\-t\-ed graph model
				has one node per pixel and contains only unary and submodular pairwise potentials.
         Without branching, SDBC converges to the exact solution, which means the relative gap between the lower- and upper-bound is smaller than $10^{-8}$.
         The assignment by SDBC (d) is consistent with that of the graph-cut algorithm (c).
         The latter is well known to provide the exact MAP solution for such binary submodular MRFs.
}}
\label{fig:denoise}
\end{figure}

\begin{table*}[tpb!]
{
  \centering
\centering
\footnotesize
\begin{tabular}{lllcccc}
\hline
Algorithm &Upper-bound &Lower-bound &$\#$Best-ub &$\#$Best-lb & $\#$Exact & Runtime\\
\hline
\multicolumn{2}{l}{{\em Runtime $\leq20$min}} & \multicolumn{5}{l}{} \\
                SDBC &$\mathbf{-504.03}$ &$\mathbf{-505.03}$ &$\mathbf{6}$ &$\mathbf{4}$ &$0$ &$1187.83$ \\
          MPLP-CP-v2 &$-502.34$ &$-514.48$ &$2$ &$2$ &$\mathbf{2}$ &$822.46$ \\
                 ILP &$-502.88$ &$-568.20$ &$1$ &$0$ &$0$ &$1220.62$ \\
       {\color{blue} LSA-TR (euc.) } &{\color{blue} $-503.87$} &{\color{blue}$-\infty$} &{\color{blue}$0$} &{\color{blue}$0$} &{\color{blue}$0$} &{\color{blue}$1200.40$} \\
       {\color{blue} LSA-TR (ham.) } &{\color{blue} $-503.76$} &{\color{blue}$-\infty$} &{\color{blue}$0$} &{\color{blue}$0$} &{\color{blue}$0$} &{\color{blue}$1200.10$} \\
                QPBO &$-501.92$ &$-\infty$ &$0$ &$0$ &$0$ &$\mathbf{0.76}$ \\
\hline
\multicolumn{2}{l}{{\em Runtime $\leq1$hr}} & \multicolumn{5}{l}{} \\
                SDBC &$\mathbf{-504.08}$ &$\mathbf{-504.45}$ &$\mathbf{6}$ &$\mathbf{4}$ &$0$ &$3460.10$ \\
          MPLP-CP-v2 &$-503.60$ &$-508.40$ &$2$ &$2$ &$\mathbf{2}$ &$\mathbf{2415.58}$ \\
                 ILP &$-502.91$ &$-566.41$ &$2$ &$0$ &$0$ &$3626.06$ \\
\hline
\multicolumn{2}{l}{{\em Runtime $\leq24$hr}} & \multicolumn{5}{l}{} \\
                SDBC &$\mathbf{-504.09}$ &$\mathbf{-504.09}$ &$\mathbf{6}$ &$\mathbf{6}$ &$\mathbf{5}$ &$\mathbf{41830.36}$ \\
          MPLP-CP-v2 &$-503.95$ &$-505.37$ &$4$ &$2$ &$2$ &$57616.95$ \\
                 ILP &$-504.07$ &$-560.11$ &$4$ &$0$ &$0$ &$86026.42$ \\
\hline
\end{tabular}
\footnotesize
\caption{
{
Image deconvolution ($6$ instances, $1219$ variables and $2$ labels for each instance).
LSA-TR achieves the second best average energy value within the runtime limit of $20$ minutes.
MPLP-CP-v2 quickly solves two easy instances \wrt $3\!\times\!3$ kernel exactly,
but becomes worse for models \wrt $5\!\times\!5$ and $7\!\times\!7$ kernel.
ILP is not able to solve any of the six instances exactly within $24$ hours,
while our method achieves the best energy values for all six instances and converges to exact solutions to five instances.}
\label{tab:deconvolution}
}}
\end{table*}

\begin{table}[tpb!]
{
  \centering
\centering
\footnotesize
\begin{tabular}{lccccc}
\hline
Kernel size &$|\setV|$ &$|\setE|$ &${|\setE|}_{sub} / |\setE|$ &$|\setV|_{redu}$\\
\hline
$3\times3$ &$1219$ & $13506$ & $0.212$ & $631.5$\\
$5\times5$ &$1219$ & $42120$ & $0.059$ & $921.5$\\
$7\times7$ &$1219$ & $82530$ & $0.030$ & $1094.5$\\
\hline
\end{tabular}
\footnotesize
\caption{
{
The graph sizes \wrt different convolution kernels.
With the growth of kernel size, the number of edges ($|\setE|$) increases and the portion of submodular edges (${|\setE|}_{sub} / |\setE|$) decreases.
There are more variables ($|\setV|_{redu}$) in the reduced graph (by QPBO) for a larger kernel size.}
\label{tab:deconvolution}
}}
\end{table}

\begin{figure}[t!]
\captionsetup[subfloat]{labelformat=empty,font=footnotesize, captionskip=1pt,nearskip=1pt,farskip=0pt,margin=0pt,topadjust=0pt}
\centering
\subfloat{
\centering
\includegraphics[type=png,ext=.png,read=.png,width=0.09\textwidth]{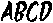}
\includegraphics[type=png,ext=.png,read=.png,width=0.09\textwidth]{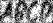}
\includegraphics[type=png,ext=.png,read=.png,width=0.09\textwidth]{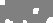}
\includegraphics[type=png,ext=.png,read=.png,width=0.09\textwidth]{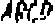}
\includegraphics[type=png,ext=.png,read=.png,width=0.09\textwidth]{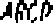}
}\\
\subfloat[{\scriptsize Ground Truth}]{
\centering \includegraphics[type=png,ext=.png,read=.png,width=0.09\textwidth]{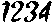}}
\subfloat[{\scriptsize Input Image}]{
\centering \includegraphics[type=png,ext=.png,read=.png,width=0.09\textwidth]{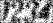}}
\subfloat[{\scriptsize QPBO}]{
\centering \includegraphics[type=png,ext=.png,read=.png,width=0.09\textwidth]{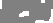}}
\subfloat[{\scriptsize MPLP-CP-v2}]{
\centering \includegraphics[type=png,ext=.png,read=.png,width=0.09\textwidth]{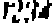}}
\subfloat[{\scriptsize SDBC}]{
\centering \includegraphics[type=png,ext=.png,read=.png,width=0.09\textwidth]{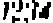}}
\caption{
{
Image deconvolution.
         The task is to recover the binary ground truth image (first column) from the blurry and noisy gray-level input image (second column).
         The third column demonstrates the partial optimal results of QPBO, in which the labelled pixels are shown in black/white color and the unlabelled pixels are shown in gray color.
         The last two columns are the images reconstructed by MPLP-CP-v2 and SDBC, using the same computation time of $20$ minutes.
         SDBC gives better energy value and recovery accuracy than MPLP-CP-v2.
}}
\label{fig:deconvolution2}
\end{figure}

\begin{figure*}[t!]
\captionsetup[subfloat]{labelformat=empty,font=footnotesize, captionskip=1pt,nearskip=1pt,farskip=0pt,margin=0pt,topadjust=0pt}
\centering
\subfloat[Ground Truth]{
\centering \includegraphics[type=png,ext=.png,read=.png,width=0.21\textwidth]{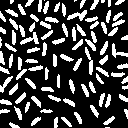}}
\subfloat[Input Image]{
\centering \includegraphics[type=png,ext=.png,read=.png,width=0.21\textwidth]{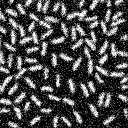}}
\subfloat[MPLP-CP-v2]{
\centering \includegraphics[type=png,ext=.png,read=.png,width=0.21\textwidth]{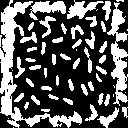}}
\subfloat[SDBC]{
\centering \includegraphics[type=png,ext=.png,read=.png,width=0.21\textwidth]{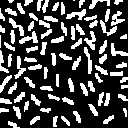}}
\caption{
{
Image deconvolution on a $128 \times 128$ image, which is convoluted by a $32 \times 32$ Gaussian kernel.
         With the same running time, the result of SDBC is much better than MPLP-CP-v2.
}
}
\label{fig:deconvolution3}
\end{figure*}

In this paper, we refer to image deconvolution as the task of reconstructing an image from its convolution with a {\em known} blurring kernel.
Raj and Zabih~\cite{raj2005graph} formulated the deconvolution to a MAP-MRF inference problem with unary and non-submodular pairwise energy functions.
The connectivity increases with the growth in kernel size.
We use two binary images and generate the deconvolution models with different kernel sizes ($3\!\times\!3$, $5\!\times\!5$ and $7\!\times\!7$).
QPBO is applied on the six deconvolution models to reduce the graph size, and then SDBC, MPLP-CP-v2 and DAOOPT are applied
on the reduced graphs.
We also report the results of QPBO and LSA-TR~\cite{gorelick2014a} with Euclidean or Hamming distance.
{ LSA-TR is repeated (around $8000$ times) with random initializations until the $20$-minute time limit is used up.}
ICM is performed as a post-processing procedure for all the evaluated methods.
Table~\ref{tab:deconvolution} shows the primary evaluation results within the runtime limits of $20$ minutes, $1$ hour and $24$ hours.
DAOOPT did not give solutions within the $24$ hours, so its result is omitted.
Within $20$ minutes,
MPLP-CP-v2 converges quickly to the exact MAP solutions for the two models with respect to $3\times3$ kernel.
But it does not converge for the rest four models with larger kernel sizes and higher connectivity.
ILP fails to solve any of the six instances exactly within $24$ hours.
LSA-TR and QPBO provide non-optimal solutions using a much shorter time.
On the other hand, SDBC gives the best upper-bounds on all the $6$ instances within any of the three time limits,
and is able to solve $5$ instances exactly (the model with respect to ``ABCD'' and $7\!\times\!7$ is not solved exactly by SDBC).

Fig.~\ref{fig:deconvolution2} illustrates the images recovered by SDBC and MPLP-CP-v2 \wrt $7\!\times\!7$ kernels within $20$ minutes.
For image ``1234'', SDBC achieves the energy value of $-478.07$ and the recovery accuracy of $87.4\%$,
which are better than those given by MPLP-CP-v2 ($-474.19$ and $85.6\%$) respectively.
Similarly for image ``ABCD'', SDBC also performs better than MPLP-CP-v2,
in terms of energy value ($-416.63$ \vs $-410.82$)
and accuracy ($87.5\%$ \vs $86.0\%$).

We also evaluate algorithms on a larger image (Fig.~\ref{fig:deconvolution3}).
The binary ground truth image is convoluted by a $32 \times 32$ Gaussian kernel.
         The corresponding MRF problem is large: $16384$ variables and $13130296$ edges,
         so we use a parallel eigen-solver,
         PLASMA\footnote{ \url{http://icl.cs.utk.edu/plasma/}}. %
         The last image shows the deconvolution result of SDBC using $5600$ seconds on $16$ cores.
         The result of MPLP-CP-v2 is demonstrated in the third image, using the same CPU hours.
         SDBC performs considerably better than MPLP-CP-v2.

\subsection{Benchmarks}

In this section, we evaluate the proposed algorithm on some MRF models in PIC2011~\cite{pic2011} and OpenGM~\cite{opengm2} benchmarks.
The experimental results show that our method performs better than state-of-the-arts on these models.
The chosen models are difficult for many conventional methods in that these inference tasks are
non-submodular,  densely connected and/or with unary terms.

\subsubsection{PIC2011-ObjectDetection}

\begin{table*}[tpb!]
{
  \centering
\centering
\footnotesize
\begin{tabular}{lllccc}
\hline
Algorithm &Upper-bound &Lower-bound &$\#$Best-ub &$\#$Best-lb & $\#$Exact \\
\hline
\multicolumn{2}{l}{{\em Runtime $=20$min}} & \multicolumn{4}{l}{} \\
                SDBC &$\mathbf{-19302.83}$ &$\mathbf{-19910.31}$ &$\mathbf{34}$ &$\mathbf{37}$ &$0$ \\
          MPLP-CP-v1 &$-18595.15$ &$-24749.11$ &$10$ &$0$ &$0$ \\
          MPLP-CP-v2 &$-18905.71$ &$-24624.95$ &$13$ &$0$ &$0$ \\
                 ILP &$-18247.73$ &$-542806.60$ &$8$ &$0$ &$0$ \\
              DAOOPT &$-19139.70$ &$-\infty$ &$17$ &$0$ &$0$ \\
             MPLP-BB &$-17838.86$ &$-36904.82$ &$11$ &$0$ &$0$ \\
\hline
\multicolumn{2}{l}{{\em Runtime $=1$hr}} & \multicolumn{4}{l}{} \\
                SDBC &$\mathbf{-19303.06}$ &$\mathbf{-19650.62}$ &$\mathbf{35}$ &$\mathbf{35}$ &$0$ \\
          MPLP-CP-v1 &$-18985.04$ &$-22312.83$ &$17$ &$1$ &$0$ \\
          MPLP-CP-v2 &$-19231.27$ &$-22260.29$ &$19$ &$1$ &$0$ \\
                 ILP &$-19049.54$ &$-34838.62$ &$15$ &$0$ &$0$ \\
              DAOOPT &$-19139.70$ &$-\infty$ &$17$ &$0$ &$0$ \\
             MPLP-BB &$-18227.13$ &$-36804.25$ &$10$ &$0$ &$0$ \\

\hline
\multicolumn{2}{l}{{\em Runtime $=24$hr}} & \multicolumn{4}{l}{} \\
     SDBC &$\mathbf{-19313.25}$ &$\mathbf{-19553.50}$ &$\mathbf{36}$ &$\mathbf{34}$ &$\mathbf{3}$ \\
          MPLP-CP-v2 &$-19270.70$ &$-21138.12$ &$22$ &$3$ &$0$ \\
                 ILP &$-19221.27$ &$-33888.65$ &$22$ &$0$ &$0$ \\
              DAOOPT &$-19139.70$ &$-\infty$ &$16$ &$0$ &$0$ \\
\hline
\end{tabular}
\footnotesize
\caption{
{
PIC2011-ObjectDetection ($37$ instances, $60$ variables and $10$-$20$ labels for each instance).
SDBC achieves the best upper-bounds and lower-bounds on most of instances within $20$ minutes, $1$ hour and $24$ hours.
Exact solutions to $3$ instances are obtained by our algorithm.
}}
\label{tab:objdet}
}
\end{table*}

There are $37$ models in the ``Object-Detection" sub-category of the PIC2011 challenge,
which are fully-connected, lacking in unary terms and have $60$ nodes and $10/15/20$ states per node.
SDBC is compared with MPLP-CP-v1, MPLP-CP-v2, ILP, DAOOPT and MPLP-BB,
among which DAOOPT is the winner of the PIC2011 challenge.

The primary results are shown in Table~\ref{tab:objdet}.
The upper-bounds and lower-bounds of different methods are compared under three running-time limits: $20$ minutes, $1$ hour and $24$ hour.
Our method achieves the best upper- and lower-bounds for most of the $37$ models under all time limits.
DAOOPT is the second best method within $20$ minites, while MPLP-CP-v2 becomes better than DAOOPT within $1$ hour.
ILP performs poorly within the runtime limit of $20$ minutes,
while it becomes the second best in terms of the number of best upper-bounds ($22$, the same as MPLP-CP-v2).

In particular, all the evaluated methods do not converge to global optimal solutions on these models,
except that SDBC gives exact solutions to three of the $37$ models.

%
%
%

\subsubsection{OpenGM-ChineseChar}

\begin{table*}[tpb!]
{
  \centering
\centering
\footnotesize
\begin{tabular}{lllcccl}
\hline
Algorithm & Upper-bound & Lower-bound &$\#$Best-ub &$\#$Best-lb & $\#$Exact & Runtime (s)\\
\hline
SDBC & $-49546.3380$ &$\mathbf{-49558.9258}$  &$\mathbf{84}$ &$\mathbf{85}$ &$44$ &$2332.1 8$\\
MCBC & $\mathbf{-49550.0972}$ &$-49612.3817$  &$72$ &$57$ &$\mathbf{56}$ &$2053.89$\\
ILP  & $-49547.4144$ &$-50061.1506$  &$59$ &$0$ &$0$ &$3553.71$\\
LSA-TR(euc.) & $-49548.0996$ &$-\infty$  &$28$ &$0$ &$0$  &$\mathbf{0.05}$\\
LSA-TR(ham.) & $-49536.7646$ &$-\infty$  &$1$ &$0$ &$0$ &$0.06$\\
QPBO & $-49501.9531$ &$-50119.3835$  &$0$ &$0$ &$0$ &$0.17$\\
\hline
\end{tabular}
\footnotesize
\caption{
{
OpenGM-ChineseChar ($100$ instances, $8000$ variables and $2$ labels for each instance).
For fair comparison, all the combinatorial methods (SDBC, MCBC and ILP) are evaluated on graphs reduced by QPBO.
The results of all algorithms except SDBC are obtained from \cite{kappes2013comparative}.
Within $1$ hour, SDBC solves $44$ instances exactly and achieves the best upper-bounds and lower-bounds on over $80\%$ instances.
}}
\label{tab:chinchar}
}
\end{table*}

The graphical model for Chinese character inpaining in the OpenGM benchmarks contains non-submoduar terms,
such that graph cuts based methods cannot give exact solutions in general.
It is shown in \cite{kappes2013comparative} that
applying combinatorial algorithms (like MCBC~\cite{bonato2014lifting} and ILP~\cite{cplex}) on the models reduced by QPBO
achieves the best performance.
Likewise, we also apply SDBC on the models reduced by QPBO and achieve comparable results.
Within one hour, our method solves $44$ out of the $100$ instances exactly,
which is only worse than the specialized solver, MCBC.
Note that MCBC utilizes more complicated tightening constraints and rounding approach than ours.
On the other hand, our method is better than all the other methods (including MCBC)
in terms of the number of best upper-bounds ($84$) and lower-bounds ($85$).
In particular, SDBC achieves better lower-bounds on those instances the LP methods (MCBC and ILP) are loose,
which demonstrate the potential of SDBC in solving these instances exactly.
To validate this potential, we run SDBC for $4$ hours and find that it gives global optimal solutions of $80$ instances.

\subsubsection{OpenGM-ModularityClustering}

\begin{table*}[tpb!]
  \centering
\centering
\footnotesize
{
\begin{tabular}{lllcccl}
\hline
Algorithm & Upper-bound & Lower-bound &$\#$Best-ub &$\#$Best-lb & $\#$Exact & Runtime (s)\\
\hline
SDBC & $\mathbf{-0.4913}$ &$\mathbf{-0.4939}$  &$\mathbf{6}$ &$\mathbf{6}$ &$\mathbf{5}$ &$1500.26$\\
KL & $-0.4860$ &$-\infty$  &$2$ &$0$ &$0$ &$\mathbf{0.01}$\\
MCR-CCFDB-OWC & $-0.4400$ &$-0.5021$  &$5$ &$5$ &$\mathbf{5}$ &$601.38$\\
MCI-CCFDB-CCIFD & $-0.4652$ &$-0.4962$  &$5$ &$5$ &$\mathbf{5}$ &$602.75$\\
MCI-CCI  & $-0.4312$ &$-0.5158$  &$4$ &$4$ &$4$ &$1207.07$\\
MCI-CCIFD &  $-0.4399$ &$-0.5176$  &$4$ &$4$ &$4$ &$1204.03$\\
\hline
\end{tabular}
\footnotesize
\caption{
{
OpenGM-ModularityClustering ($6$ instances, $34$-$115$ variables and $34$-$115$ labels for each instance).
As the top-performing algorithms (MCR-CCFDB-OWC and MCI-CCFDB-CCIFD) in \cite{kappes2013comparative},
SDBC solves $5$ instances exactly. On the remaining largest instance, our method achieves the best lower-bound
and upper-bound over all compared methods within $1$ hour.
}
}
\label{tab:moducluster}
}
\end{table*}

The graphical models in this subclass is to find a clustering of a network maximizing the modularity.
Although the six instances have small graph sizes ($34$ to $115$ variables), they are difficult
due to the absence of unary terms and fully-connected graph structure.

It can be seen in \cite{kappes2013comparative} that the LP relaxation method
with cycle and odd-wheel constraints (MCR-CCFDB-OWC) \cite{kappes2011globally} and the LP/ILP method (MCI-CCFDB-CCIFD)~\cite{kappes2013higher}
perform the best on modularity clustering, which solve five of the six instances exactly.
However, they do not converge on the largest instance within one hour.
Kerninghan-Lin (KL) algorithm~\cite{kernighan1970a},
a specialized efficient heuristic for clustering networks,
offers a better solution than MCI and MCR on this instance.
As MCR and MCI, our method also solves these five instances within one hour {\em without branching}
 (in other words, using only one bounding procedure).
Moreover, SDBC provides the best solution (in terms of upper-bound) to the largest instance among
all the evaluated approaches.

\begin{table*}[tpb!]
  \centering
\centering
\footnotesize
{
\begin{tabular}{llccccc}
\hline
Dataset &Solution &$\#$Instances &$\#$Bounding &$\#$Pruned &$|\setQ|$ &Runtime\\
\hline
{Deconvolution }    &Exact   &$5$   &$10.80$ &$5.90$ &$0.00$ &$33332$\\
($6$ instances, runtime $\leq 24$hr)                                   &Inexact &$1$   &$19.00$ &$6.00$ &$8.00$ &$84324$\\
\hline
{Object-Detection } &Exact    &$3$   &$75.67$ &$38.33$ &$0.00$  &$33625$\\
($37$ instances, runtime $\leq 24$hr)                                   &Inexact  &$34$  &$74.68$ &$19.50$ &$36.68$ &$85631$\\
\hline
{ChineseChar }   &Exact    &$80$   &$3.42$ &$2.21$ &$0.00$ &$4347$\\
($100$ instances, runtime $\leq 4$hr)                                &Inexact  &$20$   &$3.45$ &$1.00$ &$2.45$ &$12392$\\
\hline
{Modularity-Clustering } &Exact &$5$   &$1.00$ &$1.00$ &$0.00$ &$1063$\\
($6$ instances, runtime $\leq 1$hr)                                      &Inexact &$1$   &$1.00$ &$0.00$ &$2.00$ &$3686$\\
\hline
\end{tabular}
\footnotesize
\caption{
{
The average number of bounding iterations, pruned subproblems and unsolved subproblems in the queue $\setQ$ for SDBC.
The results for instances solved exactly and inexactly are summarized individually.
SDBC uses $1$-$10$ bounding iterations to reach the exact solutions to most of instances on Deconvolution, ChineseChar and Modularity-Clustering.
For Object-Detection, SDBC is able to solve $3$ instances exactly with less than $100$ bounding evaluations.
}
}
}
\end{table*}

} %

\section{Conclusion}
We have presented an efficient Branch-and-Cut method for MAP-MRF inference problems
by taking the advantage of an efficient and tight SDP bounding procedure.
The main contribution of the proposed method include:
{
1) A variety of linear tightening constraints have been incorporated in SDP relaxation to MAP problems,
by using an efficient SDP solver and cutting-plane.
2) Several techniques have also been employed to make the cutting-plane and branch-and-bound more efficient,
including model reduction, warm start and dropping inactive constraints. }
Experiments de\-monstrate that the proposed method performs very well, particularly for problems
with high connectivity or weak unary
potentials.
These types of problems are difficult to  solve by almost all other existing methods.
Furthermore, our method usually provide good approximate solutions within the first several bounding procedures.
In the experiments, we have compared the proposed approach with state-of-the-art methods.
The results demonstrate the superior performance of our approach on the evaluated problems.
%

%

\section{Appendix}

\subsection{Relationship between the Standard SDP Relaxation \eqref{eq:map} and
the Simplified Dual \eqref{eq:fastsdp_dual}}
\label{app:1}

The Lagrangian dual of \eqref{eq:map} can be expressed in the following general form:
\begin{subequations}
\label{eq:standard_sdpdual}
\begin{align}
\min_{\bu} &\quad \bu^\T \bb \\
\sst           &\quad \bZ = \bA + \textstyle{\sum_{i=1}^m} u_i \bB_i \psdd \mathbf{0}, \label{eq:standard_sdpdual_cons}\\
               &\quad u_i \geq 0, \forall i \in \setI_{in}.
\end{align}
\end{subequations}
The \psd constraint \eqref{eq:standard_sdpdual_cons} can be replaced
by a {\em penalty function}, which is considered as a measure of violation of this constraint.
In our case, the penalty function is defined as
$\mathrm{p}(\bu) = \lVert \min(\mathbf{0}, {\boldsymbol \lambda}) \rVert_2^2 = \lVert \Pi_{\mathcal{S}^{n\!h\!+\!1}_+} (\bC(\bu)) \rVert_F^2 $,
where ${\boldsymbol \lambda}$ is the vector of eigenvelues of $\bZ$.
We can find that if $\mathrm{p}(\bu) = 0$, then $\bZ \psdd \mathbf{0}$.
Now the problem \eqref{eq:standard_sdpdual} can be transformed to
\begin{subequations}
\label{eq:penalty_form}
\begin{align}
\min_{\bu} &\quad \bu^\T \bb + \frac{\gamma}{2}\lVert \Pi_{\mathcal{S}^{n\!h\!+\!1}_+} (\bC(\bu)) \rVert_F^2\\
\sst           &\quad u_i \geq 0, \forall i \in \setI_{in},
\end{align}
\end{subequations}
where $\gamma > 0$ serves as a penalty parameter. With the increase of $\gamma$,
the solution to \eqref{eq:penalty_form} converges to that of \eqref{eq:standard_sdpdual}.
It is clear that \eqref{eq:penalty_form} is equivalent to \eqref{eq:fastsdp_dual}.

\subsection{Proof of Propositions\ \ref{prop1} and \ \ref{prop2}}
\label{app:2}

Firstly, It is known \cite{Malick2007spherical,peng2013cpvr} that the set of \psd matrices with fixed trace
$\Theta_\eta := \{ \bX \psdd \mathbf{0} | \mathrm{trace}(\bX) \!=\! \eta\}$, $\forall \eta > 0$
has the following property:
\begin{theorem} (The spherical constraint).
\label{thm:1}
$\forall \eta >0, \forall \bX \in \Theta_\eta$, we have $\lVert \bX \rVert_{\!F\!} \!\leq\! \eta$,
and $\lVert \bX \rVert_{\!F\!} \!=\! \eta$ if and only if $\mathrm{rank}(\bX) \!=\! 1$.
\end{theorem}
It is also shown in \cite{peng2013cpvr} that
the problem \eqref{eq:fastsdp_dual} is the Lagrangian dual of the following problem:
\begin{subequations}
\label{eq:map_sdcut}
\begin{align}
\min_{\by,\bY} \,\, & \mathrm{E}(\by,\bY)
  + \mathrm{g}_\gamma(\by,\bY) \\
\sst \,\,
&
\eqref{eq:sdprelax_cons1},
\eqref{eq:sdprelax_cons2},
\eqref{eq:sdprelax_cons3},
\eqref{eq:sdprelax_cons4},
\eqref{eq:sdprelax_cons5},
\eqref{eq:tricons}, \eqref{eq:cyccons}, \eqref{eq:owccons}, \\
& \Omega(\by, \bY) \psdd \mathbf{0},
\end{align}
\end{subequations}
where $\mathrm{g}_\gamma(\by,\bY) = \frac{1}{2\gamma}(\lVert \Omega(\by,\bY)
\rVert^2_F - (n+1)^2)$.

\begin{proof}{{\em of} {\bf Proposition~\ref{prop1}}:}
($i$) $\forall \setD_1 \!\subseteq\! \setD_2 \!\subseteq\! \setZ^n$,
$\exists \setF_{in}, \setF_{eq} \in \{(p,i) \}_{p \in \setV, i \in \setZ}$
such that $\setD_1 \!=\! \{ \bx \!\in\! \setD_2 \ | \ x_p \!\neq\! i, \forall (p,i) \!\in\! \setF_{in};
                                              x_p \!=\! i, \forall (p,i) \!\in\! \setF_{eq} \}$.
Consequently, the difference between the SDCut primal formulation \eqref{eq:map_sdcut} \wrt $\setD_1$ and $\setD_2$
is that the one \wrt $\setD_1$ contains the following additional linear constraints:
\begin{align}
\label{eq:addi_cons_d1d2}
\left\{ \begin{array}{ll} y_{p,i} = 0, Y_{pi,qj} = Y_{qj,pi} = 0,       &\forall (p,i) \in \setF_{in}, \\
                          y_{p,i} = 1, Y_{pi,qj} = Y_{qj,pi} = y_{q,j}, &\forall (p,i) \in \setF_{eq}.
       \end{array}
\right.
\end{align}
Because of the strong duality, we know that $\mathrm{d}^\star_\gamma(\setD)$ equals to
the optimal value of the corresponding primal problem \eqref{eq:map_sdcut}.
Then we have $\mathrm{d}^\star_\gamma(\setD_1) \geq \mathrm{d}^\star_\gamma(\setD_2)$,
as the primal problem \eqref{eq:map_sdcut} \wrt $\setD_1$
has more constraints than that \wrt $\setD_2$.

\noindent
($ii$) This proof is simple. As $\lvert {\mathcal{D}} \rvert = 1$, there is only one point $\hat{\bx}$ in the set $\setD$
 and $\mathrm{E}(\hat{\bx}) = \min_{\bx \in \setD} \mathrm{E}(\bx)$.
Then the feasible set of \eqref{eq:map_sdcut} also contains a single point $\{ \hat{\by}, \hat{\bY} \}$
corresponding to $\hat{\bx}$ by applying constraints as \eqref{eq:addi_cons_d1d2}.
Because $\lVert \Omega(\hat{\by},\hat{\bY}) \rVert^2_F = (n+1)^2$,
we have $\mathrm{d}^\star_\gamma (\setD) = \mathrm{E}(\hat{\by},\hat{\bY}) = \min_{\bx \in \setD} \mathrm{E}(\bx)$.
\end{proof}

\begin{proof}{{\em of} {\bf Proposition~\ref{prop2}}:}
$\{ \by_\gamma^\star, \bY_\gamma^\star \}$ is
the optimal solution of \eqref{eq:map_sdcut} based on the strong duality, and
$\mathrm{d}_\gamma(\bu^\star_\gamma)$
is the corresponding optimal objective value.
Consider the following problem
\begin{subequations}
\label{eq:map_sdcut_rank1}
\begin{align}
\min_{\by,\bY} \,\, & \mathrm{E}(\by,\bY)
  + \mathrm{g}_\gamma(\by,\bY) \\
\sst \,\,
& \Omega(\by, \bY) \psdd \mathbf{0}, \,
\mathrm{rank}(\Omega({\by_\gamma^\star, \bY_\gamma^\star})) = 1, \\
& \eqref{eq:sdprelax_cons1},
\eqref{eq:sdprelax_cons2},
\eqref{eq:sdprelax_cons3},
\eqref{eq:sdprelax_cons4},
\eqref{eq:sdprelax_cons5},
\eqref{eq:tricons}, \eqref{eq:cyccons}, \eqref{eq:owccons}, \label{eq:map_sdcut_rank1_lcons}
\end{align}
\end{subequations}
which adds a rank-$1$ constraint to the problem~\eqref{eq:map_sdcut}.
Then
$\mathrm{d}_\gamma(\bu^\star_\gamma)$ and  $\{ \by_\gamma^\star, \bY_\gamma^\star \}$
are also optimal for the above problem.
Note that the constraints \eqref{eq:sdprelax_cons1},
\eqref{eq:sdprelax_cons2},
$\Omega({\by_\gamma^\star,
\bY_\gamma^\star}) \psdd \mathbf{0}$ and
$\mathrm{rank}(\Omega({\by_\gamma^\star, \bY_\gamma^\star})) = 1$,
 force $\{ \by_\gamma^\star, \bY_\gamma^\star \}$ to be a vertex of $\mathcal{M}(\mathcal{G},\setZ)$.
So the feasible set of \eqref{eq:map_sdcut_rank1} is
$\mathcal{M}(\mathcal{G},\setZ)$.
On the other hand, $\mathrm{g}_\gamma(\by_\gamma^\star, \bY_\gamma^\star) = 0$
at $\mathrm{rank}(\Omega({\by_\gamma^\star, \bY_\gamma^\star})) = 1$
(Theorem~\ref{thm:1}), so the objective function of
\eqref{eq:map_sdcut_rank1} is $\mathrm{E}(\by,\bY) $.
In summary,
the problem \eqref{eq:map_sdcut_rank1}
is equivalent to the MAP problem
$\displaystyle{\min_{\by,\bY \in \mathcal{M}(\mathcal{G},\setZ)}} \mathrm{E}(\by,\bY)$.
Then we have that
$\by_\gamma^\star, \bY_\gamma^\star$ yield the exact MAP solution
 and $\mathrm{d}_\gamma(\bu^\star_\gamma)$
is the minimum energy.
The value of $\gamma$ does not affect the above results.
\end{proof}

{
\bibliographystyle{IEEEtranN}
\bibliography{cs_sdp}
}

\end{document}